%% file: main.tex
\pdfoutput=1
\documentclass[11pt]{article}
\usepackage[table]{xcolor}
\usepackage[final]{acl}  
\usepackage{times}
\usepackage{latexsym}
\usepackage[T1]{fontenc}
\usepackage[utf8]{inputenc}
\usepackage{microtype}
\usepackage{inconsolata}
\usepackage{graphicx}
\usepackage{booktabs}
\usepackage{rotating}
\usepackage{arydshln}
\usepackage{multirow}
\usepackage{array}
\usepackage{makecell}
\usepackage{hyperref}
\usepackage{enumitem}
\usepackage{amssymb}
\usepackage{booktabs}

\usepackage[most]{tcolorbox}
\newtcolorbox{promptbox}[1][]{
    boxrule=0.5mm,
    arc=1mm,
    boxsep=0mm,
    fontupper=\ttfamily\scriptsize,
    width=0.45\textwidth,
    title=#1,
    fonttitle=\ttfamily\footnotesize\centering,
}

\newtcolorbox{promptboxfullwidth}[1][]{
    boxrule=0.5mm,
    arc=1mm,
    boxsep=0mm,
    fontupper=\ttfamily\scriptsize,
    width=\textwidth,
    title=#1,
    fonttitle=\ttfamily\footnotesize\centering,
}

\title{Sketch-of-Thought: Efficient LLM Reasoning \\ with Adaptive Cognitive-Inspired Sketching}

\author{
    Simon A. Aytes$^{1}$ \;\;
    Jinheon Baek$^{1}$ \;\;
    Sung Ju Hwang$^{1,2}$ \\
    KAIST$^{1}$ \;\; DeepAuto.ai$^{2}$ \\
    \texttt{\{saytes, jinheon.baek, sungju.hwang\}@kaist.ac.kr}
}

\begin{document}
\maketitle
\begin{abstract}
Recent advances in large language models (LLMs) have enabled strong reasoning capabilities through Chain-of-Thought (CoT) prompting, which elicits step-by-step problem solving, but often at the cost of excessive verbosity in intermediate outputs, leading to increased computational overhead. We propose \textit{Sketch-of-Thought} (SoT), a prompting framework that integrates cognitively inspired reasoning paradigms with linguistic constraints to reduce token usage while preserving reasoning accuracy. SoT is designed as a flexible, modular approach and is instantiated with three paradigms—\textit{Conceptual Chaining}, \textit{Chunked Symbolism}, and \textit{Expert Lexicons}—each tailored to distinct reasoning tasks and selected dynamically at test-time by a lightweight routing model. Across 18 reasoning datasets spanning multiple domains, languages, and modalities, SoT achieves token reductions of up to 84\% with minimal accuracy loss. In tasks such as mathematical and multi-hop reasoning, it even improves accuracy while shortening outputs.

\let\thefootnote\relax\footnotetext{Code: \url{https://www.github.com/SimonAytes/SoT}}

\end{abstract}

\input{sections/01_introduction}
\input{sections/02_methodology}
\input{sections/03_experiments}
\input{sections/04_results}
\input{sections/05_related_work}
\input{sections/06_conclusion}
\input{sections/07_limitations_future_work}
\input{sections/08_ethics_statement}
\input{sections/09_acknowledgments}
\bibliography{main}

\clearpage
\appendix
\input{appendices/APP_01_AdditionalInformation}
\input{appendices/APP_02_SystemPromptsExemplars}
\input{appendices/APP_04_ParadigmAssignments}
\input{appendices/APP_03_AdditionalExperiments}
\input{appendices/APP_05_IO_Examples}
\input{_tables/APPENDIX_summary_statistics_QWEN}
\input{_tables/APPENDIX_summary_statistics_LLAMA_PRIVATE}
\input{_tables/APPENDIX_Dominant_Paradigms}
\input{appendices/APP_Reasoning_IO_Examples}
\input{appendices/APP_Paradigm_PromptsExemplars}
\input{appendices/APP_ClassificationPrompt}

\end{document}

%% file: sections/01_introduction.tex
\section{Introduction}
\label{sec:introduction}

\begin{figure}[t]
  \centering
  \includegraphics[keepaspectratio, width=0.45\textwidth]{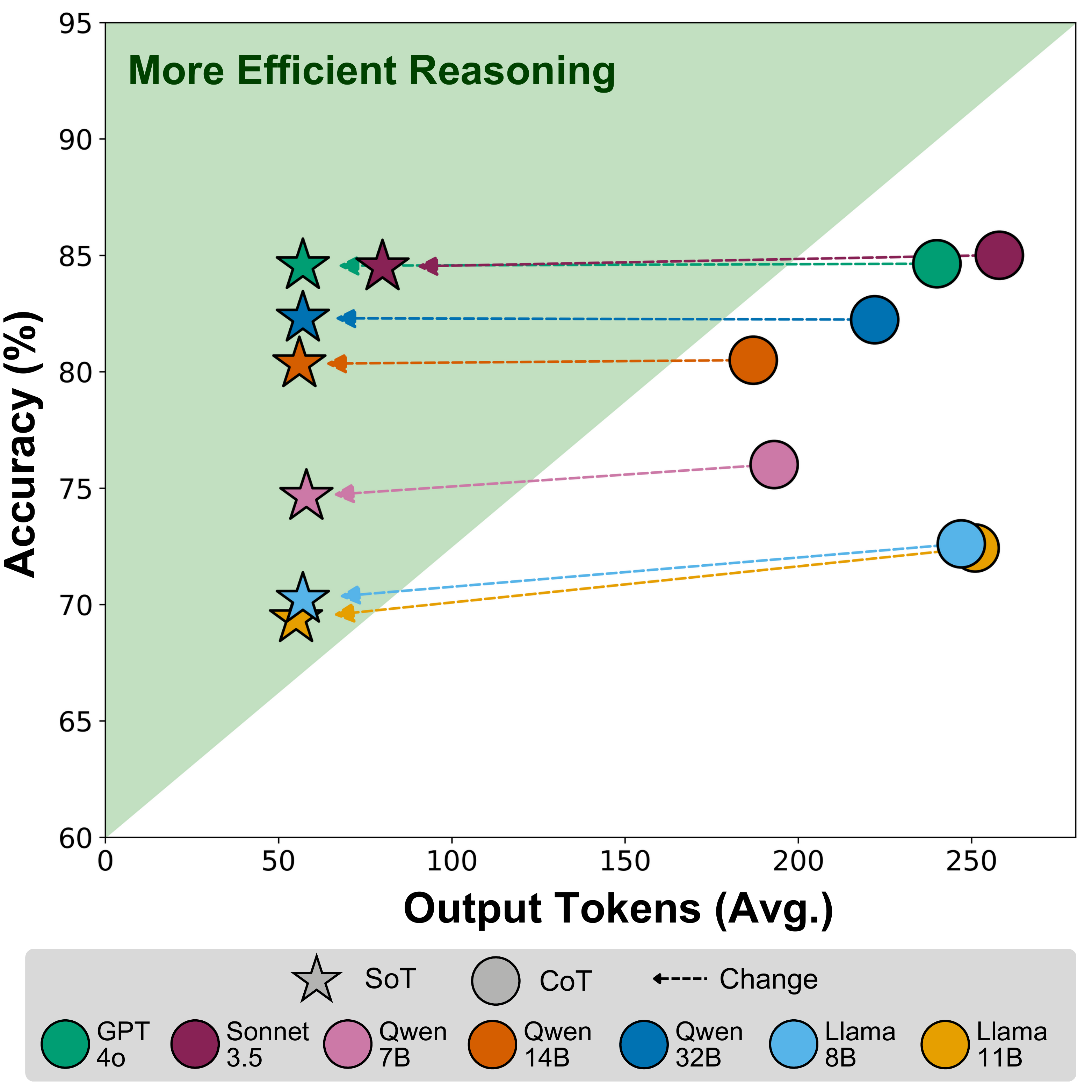}
  \vspace{-0.05in}
  \caption{A comparison of accuracy and token usage in Chain-of-Thought (CoT)~\cite{cot} and the proposed Sketch-of-Thought (SoT). Average scores for model performance across 18 datasets. Shaded region represents more efficient reasoning.}
  \label{fig:SoT_Example}
  \vspace{-0.10in}
\end{figure}

Large language models (LLMs) have become central to a wide range of complex reasoning tasks across diverse domains, such as mathematics, science, and commonsense inference~\cite{bubeck2023sparks, llmsurvey2024}. Even without dedicated training for reasoning, these models often exhibit emergent capabilities when prompted to decompose problems into intermediate steps~\cite{cot}. Chain-of-Thought (CoT) prompting~\cite{cot} exemplifies this approach by encouraging step-by-step natural language reasoning, which has been shown to significantly improve performance on tasks such as logical inference and numerical problem solving~\cite{sprague2024cotsurvey}.

Despite its benefits, CoT often produces verbose outputs that dramatically increase token usage and computational overhead, making it less suitable for latency- or budget-constrained deployment scenarios~\cite{ccot-45, arora2025training}. More sophisticated strategies, such as Self-Consistency~\cite{self-consistency}, Tree-of-Thoughts~\cite{tot}, and Graph-of-Thoughts~\cite{got}, further expand the reasoning process via structured exploration, but tend to exacerbate inefficiencies in token usage.

To tackle these limitations, we introduce \textit{Sketch-of-Thought} (SoT), a prompting framework that rethinks how language models externalize reasoning. Inspired by cognitive science, particularly the use of symbolic \textit{sketches} as efficient mental intermediaries~\cite{sketchesofthought}, SoT guides models to produce concise, structured reasoning steps that capture essential logic while avoiding full-sentence elaboration. These representations are analogous to mathematical notation or expert shorthand, preserving semantic fidelity while minimizing redundancy.

To implement this framework, we define three cognitively motivated reasoning paradigms: \textit{Conceptual Chaining}, based on associative memory; \textit{Chunked Symbolism}, grounded in working memory theory; and \textit{Expert Lexicons}, inspired by domain-specific schemas used by specialists. Each paradigm is designed for a distinct class of reasoning tasks and is implemented using training-free prompts. To support adaptive paradigm selection, we incorporate a lightweight routing model that analyzes query structure to determine the most suitable reasoning style at inference time.

We extensively evaluate SoT on 18 reasoning datasets spanning mathematical, commonsense, logical, multi-hop, scientific, and medical domains. Experimental results show that SoT reduces output token usage by up to 84\% compared to traditional CoT prompting, with no significant loss in accuracy—and even improving performance in some domains. Additional multilingual and multimodal evaluations demonstrate SoT’s ability to generalize across both languages and input modalities.

\noindent Our key contributions are as follows:

\vspace{-0.075in}
\begin{itemize}
    \item We introduce \textit{Sketch-of-Thought} (SoT), a prompting framework that leverages cognitively inspired reasoning paradigms to produce concise and structured model outputs.
    \item We present a lightweight routing model that dynamically selects the optimal reasoning paradigm based on the input query’s structure and semantics.
    \item On a battery of tests, we show that SoT significantly reduces token usage while maintaining or improving accuracy across diverse datasets, models, languages, and modalities.
\end{itemize}

%% file: sections/02_methodology.tex
\begin{figure*}[!]
    \centering
    \includegraphics[width=0.975\linewidth]{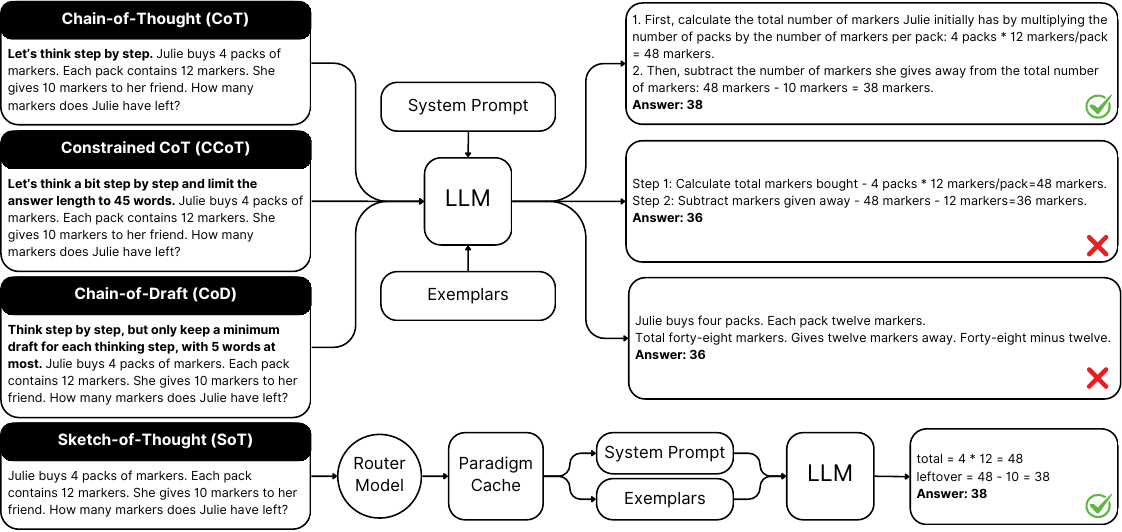}
    \vspace{-0.05in}
    \caption{\textbf{Illustration of reasoning workflows}, including the input format, intermediate reasoning structure, and output style, across four prompting methods: Chain-of-Thought (CoT)~\cite{cot}, Constrained CoT (CCoT)~\cite{ccot-45}, Chain-of-Draft (CoD)~\cite{cod}, and Sketch-of-Thought (SoT). While CoT produces verbose natural language traces, CCoT and CoD apply explicit constraints on reasoning length. SoT introduces paradigm-guided sketching, yielding more compact yet structured intermediate steps via dynamic routing.}
    \label{fig:sot-sot-flow}
    \vspace{-0.1in}
\end{figure*}

\section{Method}
\label{sec:sot}

This section outlines the technical implementation of \textit{Sketch-of-Thought} (SoT), a framework designed to improve the efficiency of reasoning in large language models while preserving performance. The end-to-end framework is illustrated in Figure~\ref{fig:sot-sot-flow}.

\subsection{Preliminary}
We first provide a brief overview of large language models (LLMs) and their application to reasoning.

\paragraph{Large Language Models}
Large language models (LLMs) are trained on massive text corpora to predict the next token in a sequence. Formally, an LLM with parameters $\theta$ receives an input token sequence $x$ and produces an output sequence $y$, denoted as follows: $y = \text{LLM}_\theta(x)$. While LLMs exhibit strong performance across a variety of domains, their computational cost increases with the number of tokens processed and generated, making efficiency a critical concern.

\paragraph{Reasoning via Prompting}
Recent LLMs exhibit emergent reasoning capabilities—the ability to solve multi-step problems without task-specific training—when guided by carefully designed prompts. Letting $P$ denote a prompt with instructions, exemplars, and the input query $q$, the process can be formalized as $r = \text{LLM}_\theta(P(q))$, where $r$ includes both the reasoning trace $[s_1, s_2, \ldots, s_m]$ and final answer $a$, as follows: $r = [s_1, s_2, \ldots, s_m, a]$. Chain-of-Thought (CoT)~\cite{cot} encourages decomposition into natural language steps.

\begin{center}
\begin{promptbox}[Traditional CoT]
Q: If a train travels 60 miles per hour for 3 hours, how far does it go?
\newline
A: <think>
1. I understand we need to find the total distance traveled by: A train moving at 60 miles per hour for a duration of 3 hours.\newline
2. To calculate the distance, I'll use the formula:\newline
Distance = Speed × Time\newline
Distance = 60 miles/hour × 3 hours\newline
3. Now I'll perform the calculation:\newline
Distance = 60 × 3 = 180 miles\newline
4. Verification:\newline
This makes sense because the train moves 60 miles each hour. After 3 hours, it will have covered 3 times that distance. </think>
\newline
Answer: 180 miles
\end{promptbox}
\end{center}

More advanced variants, for example, Tree-of-Thoughts and Graph-of-Thoughts~\cite{tot, got}, extend this idea by exploring alternative structural formulations. Despite their stylistic differences (leading to performance gains), all CoT-style prompting methods share a common limitation: increased token usage and reduced efficiency, when compared to standard direct-answer prompting \cite{arora2025training}.

\subsection{Sketch-of-Thought (SoT)}
\label{sec:paradigms}

Sketch-of-Thought (SoT) mitigates reasoning inefficiency by restructuring how models express intermediate steps. Unlike prior methods that reduce prompt length via input compression~\cite{jiang2023llmlingua, huang2024fewer}, SoT compresses the reasoning process using cognitively inspired prompts that elicit concise, structured steps.

Formally, we define different paradigm-specific prompt templates $P_{\text{SoT}}$, which steer the model to produce sketched reasoning: $[\hat{s}_1, \hat{s}_2, \ldots, \hat{s}_m, a] = \text{LLM}_\theta(P_{\text{SoT}}(q))$, where each $\hat{s}_i$ conveys the same logical content as $s_i$ (from CoT, for example), but using significantly fewer tokens, i.e., $|\hat{s}| < |s|$. These prompts enforce both linguistic constraints and cognitive structuring tailored to the task type.

As an initial realization of SoT, we create three reasoning paradigms inspired by cognitive science, each designed to align with distinct patterns found across a range of reasoning tasks.

\paragraph{Conceptual Chaining.} 
Rooted in cognitive science principles of how humans connect and retrieve related information, this paradigm creates concise logical sequences between key concepts. It draws from episodic buffer integration \cite{baddeley_episodic_2000}, the cognitive mechanism that temporarily holds and links information from different sources, and associative memory networks \cite{anderson_spreading_1983}, which describe how activating one concept automatically triggers related concepts in our minds (like how thinking of "rain" might immediately evoke "umbrella"). \textit{Conceptual Chaining} extracts essential terms and presents reasoning as direct step-by-step pathways with minimal text.

\begin{center}
\begin{promptbox}[Conceptual Chaining]
Q: What is the name of the currency used in Seoul?
\newline
A: <think>
\#Seoul → \#South Korea → Won
</think>
\newline
Answer: Korean Won
\end{promptbox}
\end{center}

\textit{Conceptual Chaining} is particularly effective for commonsense, multi-hop, logical, and scientific reasoning tasks, where establishing structured relationships between ideas is critical.

\paragraph{Chunked Symbolism.} 
Based on working memory chunking theory \cite{miller_magical_1956}, this paradigm organizes numerical and symbolic reasoning into compact, structured steps. This seminal cognitive science research showed that humans can only hold about 7$\pm$2 (i.e., 5 to 9) distinct items in working memory at once, but we overcome this limitation by "chunking" related information into meaningful units—like remembering phone numbers as area code, prefix, and line number instead of 10 separate digits. \textit{Chunked Symbolism} applies this principle by condensing mathematical reasoning into dense symbolic representations that pack more information into fewer tokens. It systematically extracts variables and performs operations while eliminating verbose explanations, using symbolic variables to transform natural language into a structured shorthand that preserves logical flow.

\begin{center}
\begin{promptbox}[Chunked Symbolism]
Q: A car accelerates at 2.5 m/s\^{}2 for 10 seconds. If its initial velocity was 15 m/s, what is its final velocity?
\newline
A: <think>
a = 2.5 m/s\^{}2, t = 10 s, vi = 15 m/s
vf = 15 + (2.5 × 10), vf = 40 m/s
</think>
\newline
Answer: 40 m/s
\end{promptbox}
\end{center}

\textit{Chunked Symbolism} excels in mathematical and arithmetic reasoning problems, where symbolic notation naturally compresses complex concepts.

\paragraph{Expert Lexicons.} 
Inspired by expert schema research \cite{chi_categorization_1981}, this paradigm leverages domain-specific shorthand and specialized notation to condense reasoning. This research demonstrated that experts in any field organize knowledge differently than novices—they develop mental frameworks (schemas) that allow them to quickly recognize patterns and use specialized terminology to communicate efficiently with peers. For example, a physician can convey complex medical conditions with a few acronyms that would require paragraphs of explanation for non-specialists. \textit{Expert Lexicons} mimics this cognitive efficiency by employing domain-specific abbreviations, notation, and symbols that pack multiple concepts into single tokens. The example below demonstrates how domain-specialized reasoning can be compressed into concise notation while preserving the critical logical connections.

\begin{center}
\begin{promptbox}[Expert Lexicons]
Q: A patient with STEMI is given MONA therapy. They are allergic to aspirin. Are they at risk with this treatment?
\newline
A: <think>
STEMI → ST-Elevation MI, MONA → {Morphine, O2, Nitrates, Aspirin}, so Aspirin $\in$ MONA
</think>
\newline
Answer: Yes
\end{promptbox}
\end{center}

\textit{Expert Lexicons} is particularly suited for technical disciplines, specialized reasoning tasks, and scenarios, where domain expertise enables significant information compression.

\subsection{Adaptive Paradigm Selection}
\label{sec:paradigm_selection}

While manual selection among three paradigms is possible for each query based on heuristic rules, such an approach is impractical at scale. Instead, we introduce a lightweight routing model that selects the paradigm dynamically based on semantic and structural features of the input query.

Given a query $q$, the routing process is denoted as follows: $P_{\text{SoT}} = \texttt{ROUTER}(q)$, where $P_{\text{SoT}}$ refers to the selected paradigm's prompt-exemplar pair and $\texttt{ROUTER}$ denotes the router model. We use DistilBERT~\cite{distilbert} as the base model due to its strong performance-efficiency trade-off and minimal inference overhead (see Appendix ~\ref{sub:router_architecture_ablation}).

\paragraph{Router Training}

We train the router model using $14{,}200$ machine-labeled examples drawn from the training splits of the datasets outlined in Section~\ref{sec:datasets}. Each sample is labeled using GPT-4o~\cite{openai2024gpt4ocard}, guided by a classification prompt derived from the paradigm definitions in Section~\ref{sec:paradigms}. We provide this classification prompt in Appendix~\ref{sub:classification_prompt}. Additionally, we evaluate GPT-4o's paradigm labeling performance in Appendix~\ref{sub:gpt4o_human_evaluation}.

To avoid overwhelming the router with irrelevant input, we replace any long or non-textual context (e.g., images or documents) with a special placeholder token (e.g., \texttt{[CONTEXT HERE]}). This ensures that the model focuses solely on the question itself, which typically contains sufficient cues for determining the appropriate reasoning style.

%% file: sections/03_experiments.tex
\section{Experimental Setup}
\label{sec:experiments}

\vspace{-0.01in}
\subsection{Datasets}
\label{sec:datasets}
\vspace{-0.02in}

To ensure a comprehensive evaluation, we validate Sketch-of-Thought (SoT) across 15 datasets spanning six categories of reasoning, following the taxonomy introduced by \citet{reasoning_task_types}. The datasets are as follows: \textbf{Mathematical Reasoning} includes GSM8K, SVAMP, AQUA-RAT, and DROP~\cite{ds_gsm8k, ds_svamp, ds_aqua_rat, ds_drop}; \textbf{Commonsense Reasoning} includes CommonsenseQA, OpenbookQA, and StrategyQA~\cite{ds_commonsenseqa, ds_openbookqa, ds_strategyqa}; \textbf{Logical Reasoning} includes LogiQA and ReClor~\cite{ds_logiqa, ds_reclor}; \textbf{Multi-Hop Reasoning} includes HotPotQA and MuSiQue-Ans~\cite{ds_hotpotqa, ds_musique_ans}; \textbf{Scientific Reasoning} includes QASC and Worldtree~\cite{ds_qasc, ds_worldtree}; and \textbf{Medical Reasoning} includes PubMedQA and MedQA~\cite{ds_pubmedqa, ds_medqa}. 

Beyond English textual reasoning, we include two additional evaluation tracks: a multilingual experiment using MMLU and its professionally translated variant MMMLU~\cite{ds_mmlu}, and a multimodal experiment using GQA \cite{ds_gqa} and the image-based subset of ScienceQA~\cite{ds_scienceqa}. Further details regarding the datasets are provided in Appendix~\ref{ap_datasets}.

\input{_tables/table_main_results}

\vspace{-0.01in}
\subsection{Baselines}
\vspace{-0.02in}

We mainly compare SoT against three established prompting-based reasoning strategies. Chain-of-Thought (CoT)~\cite{cot} elicits step-by-step natural language reasoning. Constrained CoT (CCoT)~\cite{ccot-45} introduces a global verbosity constraint, limiting the total reasoning chain to a fixed number of words—in our case, 45 words (CCoT-45). Chain-of-Draft (CoD)~\cite{cod} adopts a similar compression strategy but imposes constraints at the step level, requiring each intermediate step be no longer than five words.

\vspace{-0.01in}
\subsection{Implementation Details}
\label{sub:implementation_details}
\vspace{-0.02in}

A diverse set of instruction-tuned LLMs is selected, spanning both open-weight and proprietary offerings. These include Qwen-2.5 in 7B, 14B, and 32B variants~\cite{qwen2.5}, LLaMA-3.1-8B~\cite{llama-3.1}, LLaMA-3.2-11B~\cite{llama-3.2}, GPT-4o~\cite{openai2024gpt4ocard}, and Claude Sonnet 3.5~\cite{claude-3.5}. For experiments involving multimodal inputs, we use Qwen-2.5-VL-7B~\cite{qwen2.5-VL}, which supports visual input processing. Unless otherwise specified, Qwen-2.5-32B serves as the default model for all other experiments. We use a temperature value of 0.5 for all models to balance output stability and diversity. For open-source models, inference is accelerated using FlashAttention2~\cite{flashattention2}. We sample 150 questions from each dataset for the sake of computational costs, and report the averaged performance over three independent runs per question. For the router model, we fine-tune DistilBERT with cross-entropy loss over 5 epochs, using a batch size of 64 and a learning rate of $2\text{e}^{-5}$. During inference, the router processes the core input query. Following previous work, we use few-shot prompting to illustrate the required reasoning style, with exemplars being generated by prompting Qwen-2.5-32B with the method-specific prompt and selecting high-quality outputs. Further information regarding prompts and exemplars can be found in Appendix~\ref{sec:appendix_prompts}.

\vspace{-0.01in}
\subsection{Evaluation Protocol}
\label{sub:evaluation_protocol}
\vspace{-0.02in}

We evaluate using two primary metrics: accuracy and output token count. For multiple-choice, yes/no, or numeric tasks, accuracy is computed via exact match with the ground truth. For open-ended generation, we follow the LLM-as-a-judge paradigm~\cite{geval}, using GPT-4o~\cite{openai2024gpt4ocard} to assess correctness. Answers are extracted according to the output format (see Appendix~\ref{sub:output_format}). We analyze efficiency through the total number of generated tokens in the intermediate reasoning.

%% file: _tables/table_main_results.tex
\begin{table*}[t]
\centering
\scriptsize
\renewcommand{\arraystretch}{0.75}
\setlength{\tabcolsep}{3.0pt}
\caption{\textbf{Main Experimental Results.} Results are shown for Sketch-of-Thought (SoT), Chain-of-Thought (CoT) \cite{cot}, Constrained Chain-of-Thought (CCoT) \cite{ccot-45}, and Chain-of-Draft (CoD) \cite{cod}. Results are grouped by reasoning type, with each entry representing the average over all associated datasets. "Acc" denotes accuracy and "Tkn" denotes the number of output tokens. In the Overall section, we report two additional metrics: the token reduction percentage (shown as "Red.") and the change in accuracy between CoT and the baseline (shown as "$\Delta$"). The best results are in \textbf{bold} and the second-best are \underline{underlined}.}
\vspace{-0.125in}
\resizebox{\textwidth}{!}{
\begin{tabular}{@{}c@{\hspace{4pt}}lcccccccccccccccc}
\toprule

\multicolumn{18}{c}{\textbf{Reasoning Task}} \\

\midrule
& & \multicolumn{2}{c}{\textbf{Mathematical}} & \multicolumn{2}{c}{\textbf{Commonsense}} & \multicolumn{2}{c}{\textbf{Logical}} & \multicolumn{2}{c}{\textbf{Multi-Hop}} & \multicolumn{2}{c}{\textbf{Scientific}} & \multicolumn{2}{c}{\textbf{Medical}} & \multicolumn{4}{c}{\textbf{Overall}} \\
\cmidrule(l{2pt}r{2pt}){3-4}
\cmidrule(l{2pt}r{2pt}){5-6}
\cmidrule(l{2pt}r{2pt}){7-8}
\cmidrule(l{2pt}r{2pt}){9-10}
\cmidrule(l{2pt}r{2pt}){11-12}
\cmidrule(l{2pt}r{2pt}){13-14}
\cmidrule(l{2pt}r{2pt}){15-18}
& \textbf{Method} & \multicolumn{1}{c}{Acc} & \multicolumn{1}{c}{Tkn} & \multicolumn{1}{c}{Acc} & \multicolumn{1}{c}{Tkn} & \multicolumn{1}{c}{Acc} & \multicolumn{1}{c}{Tkn} &
\multicolumn{1}{c}{Acc} & \multicolumn{1}{c}{Tkn} & \multicolumn{1}{c}{Acc} & \multicolumn{1}{c}{Tkn} & \multicolumn{1}{c}{Acc} & \multicolumn{1}{c}{Tkn}
& \multicolumn{1}{c}{Acc $\uparrow$} & \multicolumn{1}{c}{Tkn $\downarrow$} & \multicolumn{1}{c}{Red.$\uparrow$} & \multicolumn{1}{c}{$\Delta$$\uparrow$} \\

\midrule
\multirow{4}{*}{\rotatebox[origin=c]{90}{\textbf{\makecell{Qwen\\2.5-32B}}}} 
& CoT & 84.17 & 222 & 91.48 & 177 & 71.23 & 298 & 79.44 & 155 & 92.89 & 213 & 67.66 & 292 & 82.24 & 222 & -- & -- \\
& CoD & 71.94 & 53 & 89.48 & 38 & 72.89 & 45 & 80.00 & 41 & 90.00 & 42 & 58.89 & 47 & 77.32 & 45 & \textbf{79.75} & -4.92 \\
& CCoT & 80.50 & 76 & 88.82 & 49 & 72.78 & 60 & 80.11 & 54 & 88.89 & 49 & 57.66 & 65 & 79.16 & 61 & 72.56 & \underline{-3.08} \\
\noalign{\vskip 0.25ex}\cdashline{2-18}\noalign{\vskip 0.50ex}
& \cellcolor{blue!5}SoT & \cellcolor{blue!5}86.94 & \cellcolor{blue!5}88 & \cellcolor{blue!5}92.00 & \cellcolor{blue!5}34 & \cellcolor{blue!5}71.00 & \cellcolor{blue!5}66 & \cellcolor{blue!5}81.89 & \cellcolor{blue!5}43 & \cellcolor{blue!5}91.34 & \cellcolor{blue!5}31 & \cellcolor{blue!5}61.11 & \cellcolor{blue!5}63 & \cellcolor{blue!5}82.30 & \cellcolor{blue!5}57 & \cellcolor{blue!5}\underline{74.36} & \cellcolor{blue!5}\textbf{0.06} \\

\midrule
\multirow{4}{*}{\rotatebox[origin=c]{90}{\textbf{\makecell{Qwen\\2.5-14B}}}} 
& CoT & 83.00 & 190 & 91.41 & 150 & 67.00 & 248 & 77.67 & 149 & 90.89 & 164 & 65.11 & 234 & 80.50 & 187 & -- & -- \\
& CoD & 69.22 & 63 & 89.04 & 41 & 66.22 & 47 & 80.44 & 46 & 89.44 & 43 & 59.00 & 52 & 75.61 & 50 & \textbf{73.23} & -4.89 \\
& CCoT & 81.33 & 115 & 90.52 & 58 & 70.00 & 89 & 78.89 & 91 & 89.44 & 55 & 61.44 & 86 & 79.76 & 85 & 54.49 & \underline{-0.74} \\
\noalign{\vskip 0.25ex}\cdashline{2-18}\noalign{\vskip 0.50ex}
& \cellcolor{blue!5}SoT & \cellcolor{blue!5}82.72 & \cellcolor{blue!5}78 & \cellcolor{blue!5}90.89 & \cellcolor{blue!5}37 & \cellcolor{blue!5}67.44 & \cellcolor{blue!5}63 & \cellcolor{blue!5}79.89 & \cellcolor{blue!5}45 & \cellcolor{blue!5}90.89 & \cellcolor{blue!5}37 & \cellcolor{blue!5}62.56 & \cellcolor{blue!5}63 & \cellcolor{blue!5}80.34 & \cellcolor{blue!5}56 & \cellcolor{blue!5}\underline{70.02} & \cellcolor{blue!5}\textbf{-0.16} \\

\midrule
\multirow{4}{*}{\rotatebox[origin=c]{90}{\textbf{\makecell{Qwen\\2.5-7B}}}}
& CoT & 77.94 & 186 & 86.52 & 158 & 65.67 & 284 & 73.22 & 137 & 88.67 & 181 & 57.11 & 249 & 74.86 & 199 & -- & -- \\
& CoD & 66.83 & 57 & 84.74 & 37 & 64.33 & 49 & 76.11 & 43 & 87.00 & 39 & 55.89 & 48 & 72.48 & 46 & \textbf{76.88} & \underline{-2.38} \\
& CCoT & 78.00 & 81 & 64.15 & 45 & 63.67 & 63 & 78.89 & 53 & 62.78 & 44 & 50.33 & 61 & 66.30 & 58 & 70.85 & -8.56 \\
\noalign{\vskip 0.25ex}\cdashline{2-18}\noalign{\vskip 0.50ex}
& \cellcolor{blue!5}SoT & \cellcolor{blue!5}79.28 & \cellcolor{blue!5}72 & \cellcolor{blue!5}86.74 & \cellcolor{blue!5}30 & \cellcolor{blue!5}62.00 & \cellcolor{blue!5}66 & \cellcolor{blue!5}76.11 & \cellcolor{blue!5}45 & \cellcolor{blue!5}87.22 & \cellcolor{blue!5}27 & \cellcolor{blue!5}53.78 & \cellcolor{blue!5}107 & \cellcolor{blue!5}74.19 & \cellcolor{blue!5}58 & \cellcolor{blue!5}\underline{70.85} & \cellcolor{blue!5}\textbf{-0.67} \\

\midrule
\multirow{4}{*}{\rotatebox[origin=c]{90}{\textbf{\makecell{Llama\\3.1-8B}}}} 
& CoT & 72.56 & 235 & 81.92 & 209 & 51.22 & 292 & 74.56 & 193 & 85.78 & 260 & 65.00 & 323 & 72.61 & 247 & -- & -- \\
& CoD & 55.28 & 73 & 80.67 & 45 & 47.22 & 58 & 73.22 & 49 & 81.00 & 47 & 66.22 & 55 & 66.56 & 56 & \textbf{77.31} & -6.05 \\
& CCoT & 65.22 & 88 & 80.89 & 58 & 51.00 & 73 & 75.45 & 60 & 85.00 & 57 & 68.11 & 73 & 70.84 & 70 & 71.64 & \textbf{-1.77} \\
\noalign{\vskip 0.25ex}\cdashline{2-18}\noalign{\vskip 0.50ex}
& \cellcolor{blue!5}SoT & \cellcolor{blue!5}64.67 & \cellcolor{blue!5}78 & \cellcolor{blue!5}81.41 & \cellcolor{blue!5}36 & \cellcolor{blue!5}48.11 & \cellcolor{blue!5}71 & \cellcolor{blue!5}77.11 & \cellcolor{blue!5}44 & \cellcolor{blue!5}83.56 & \cellcolor{blue!5}35 & \cellcolor{blue!5}66.44 & \cellcolor{blue!5}63 & \cellcolor{blue!5}70.22 & \cellcolor{blue!5}57 & \cellcolor{blue!5}\underline{76.91} & \cellcolor{blue!5}\underline{-2.39} \\

\midrule
\multirow{4}{*}{\rotatebox[origin=c]{90}{\textbf{\makecell{Llama\\3.2-11B}}}} 
& CoT & 70.55 & 232 & 82.74 & 216 & 50.33 & 297 & 73.45 & 198 & 85.78 & 263 & 68.44 & 334 & 72.43 & 251 & -- & -- \\
& CoD & 56.17 & 67 & 80.89 & 43 & 48.22 & 51 & 74.00 & 46 & 79.44 & 44 & 65.00 & 50 & 66.71 & 52 & \textbf{79.25} & -5.72 \\
& CCoT & 64.56 & 79 & 80.81 & 59 & 51.89 & 69 & 73.00 & 62 & 84.22 & 57 & 68.34 & 71 & 70.37 & 67 & 73.27 & \textbf{-2.06} \\
\noalign{\vskip 0.25ex}\cdashline{2-18}\noalign{\vskip 0.50ex}
& \cellcolor{blue!5}SoT & \cellcolor{blue!5}64.50 & \cellcolor{blue!5}75 & \cellcolor{blue!5}81.48 & \cellcolor{blue!5}35 & \cellcolor{blue!5}45.34 & \cellcolor{blue!5}69 & \cellcolor{blue!5}77.89 & \cellcolor{blue!5}44 & \cellcolor{blue!5}79.44 & \cellcolor{blue!5}36 & \cellcolor{blue!5}66.56 & \cellcolor{blue!5}64 & \cellcolor{blue!5}69.39 & \cellcolor{blue!5}55 & \cellcolor{blue!5}\underline{78.06} & \cellcolor{blue!5}\underline{-3.04} \\

\midrule
\multirow{4}{*}{\rotatebox[origin=c]{90}{\textbf{GPT-4o}}} 
& CoT & 85.44 & 240 & 92.74 & 200 & 74.78 & 311 & 81.56 & 156 & 93.22 & 240 & 75.22 & 308 & 84.64 & 240 & -- & -- \\
& CoD & 83.17 & 71 & 87.11 & 50 & 71.56 & 62 & 82.56 & 53 & 90.67 & 55 & 46.33 & 63 & 78.41 & 60 & \underline{74.95} & -6.23 \\
& CCoT & 83.72 & 93 & 90.59 & 63 & 71.22 & 69 & 82.33 & 70 & 90.22 & 63 & 56.22 & 71 & 80.44 & 74 & 69.11 & \underline{-4.20} \\
\noalign{\vskip 0.25ex}\cdashline{2-18}\noalign{\vskip 0.50ex}
& \cellcolor{blue!5}SoT & \cellcolor{blue!5}86.17 & \cellcolor{blue!5}69 & \cellcolor{blue!5}92.52 & \cellcolor{blue!5}39 & \cellcolor{blue!5}73.22 & \cellcolor{blue!5}80 & \cellcolor{blue!5}84.78 & \cellcolor{blue!5}47 & \cellcolor{blue!5}92.56 & \cellcolor{blue!5}39 & \cellcolor{blue!5}72.44 & \cellcolor{blue!5}61 & \cellcolor{blue!5}84.55 & \cellcolor{blue!5}57 & \cellcolor{blue!5}\textbf{76.20} & \cellcolor{blue!5}\textbf{-0.09} \\

\midrule
\multirow{4}{*}{\rotatebox[origin=c]{90}{\textbf{\makecell{Claude\\Sonnet 3.5}}}} 
& CoT & 87.11 & 233 & 91.26 & 242 & 75.22 & 314 & 81.67 & 206 & 93.89 & 264 & 75.67 & 321 & 85.01 & 258 & -- & -- \\
& CoD & 82.00 & 78 & 91.33 & 61 & 75.78 & 96 & 82.00 & 63 & 91.33 & 67 & 76.22 & 105 & 83.51 & 77 & \textbf{70.16} & \underline{-1.50} \\
& CCoT & 82.94 & 97 & 72.44 & 80 & 64.67 & 91 & 80.89 & 85 & 68.33 & 83 & 55.78 & 103 & 72.56 & 90 & 65.12 & -12.45 \\
\noalign{\vskip 0.25ex}\cdashline{2-18}\noalign{\vskip 0.50ex}
& \cellcolor{blue!5}SoT & \cellcolor{blue!5}84.06 & \cellcolor{blue!5}85 & \cellcolor{blue!5}91.11 & \cellcolor{blue!5}59 & \cellcolor{blue!5}75.00 & \cellcolor{blue!5}112 & \cellcolor{blue!5}84.44 & \cellcolor{blue!5}57 & \cellcolor{blue!5}91.78 & \cellcolor{blue!5}62 & \cellcolor{blue!5}77.78 & \cellcolor{blue!5}116 & \cellcolor{blue!5}84.50 & \cellcolor{blue!5}80 & \cellcolor{blue!5}\underline{68.99} & \cellcolor{blue!5}\textbf{-0.51} \\
[0.50ex]

\midrule
\multirow{4}{*}{\rotatebox[origin=c]{90}{\textbf{\makecell{All\\Models}}}}
& CoT & 80.11 & 220 & 88.30 & 193 & 65.06 & 292 & 77.37 & 171 & 90.16 & 226 & 67.75 & 294 & 78.12 & 233 & -- & -- \\
& CoD & 69.23 & 66 & 86.18 & 45 & 63.75 & 58 & 78.33 & 49 & 86.98 & 48 & 61.08 & 60 & 74.26 & 54 & \textbf{76.82} & \underline{-3.86} \\
& CCoT & 76.61 & 90 & 81.17 & 59 & 63.60 & 73 & 78.51 & 68 & 81.27 & 58 & 59.70 & 76 & 73.48 & 71 & 69.53 & -4.64 \\
\noalign{\vskip 0.25ex}\cdashline{2-18}\noalign{\vskip 0.50ex}
& \cellcolor{blue!5}SoT & \cellcolor{blue!5}78.33 & \cellcolor{blue!5}78 & \cellcolor{blue!5}88.02 & \cellcolor{blue!5}39 & \cellcolor{blue!5}63.16 & \cellcolor{blue!5}75 & \cellcolor{blue!5}80.30 & \cellcolor{blue!5}46 & \cellcolor{blue!5}88.11 & \cellcolor{blue!5}38 & \cellcolor{blue!5}65.81 & \cellcolor{blue!5}77 & \cellcolor{blue!5}77.29 & \cellcolor{blue!5}59 & \cellcolor{blue!5}\underline{74.68} & \cellcolor{blue!5}\textbf{-0.83} \\

\bottomrule
\end{tabular}
}
\vspace{-0.1in}
\label{tab:primary_res}
\end{table*}

%% file: sections/04_results.tex
\section{Results and Discussion}
\label{sec:results}

\subsection{Overall Performance}

As shown in Table~\ref{tab:primary_res}, Sketch-of-Thought (SoT) consistently reduces output token count while minimizing the impact on reasoning accuracy across all evaluated models. On average, SoT achieves a token reduction of over 74\% relative to CoT, with accuracy deviations typically within 1\%. These trends hold across both open-weight models and proprietary models, confirming SoT’s generalizability across architectures and model families. SoT also demonstrates strong stability across reasoning tasks, consistently balancing token reduction with minimal accuracy variance, unlike other baselines which exhibit greater fluctuations. Notably, across all runs, we found that SoT consistently reduces token usage while having a statistically insignificant impact on accuracy ($p<0.05$).

\subsection{Model-wise Trends}

Performance gains with SoT are especially notable in the Qwen family of models. On Qwen-2.5-32B, SoT achieves 82.30\% accuracy—slightly above CoT’s 82.24\%—while reducing output token count by 74.36\%. Similar patterns hold at the 14B and 7B scales, where SoT maintains accuracy within 1\% of CoT while reducing output length by over 70\%. On GPT-4o, SoT achieves 84.55\% accuracy—just 0.09\% below CoT—while reducing token usage by 76\%. Claude Sonnet 3.5 shows similar behavior, with SoT reaching 84.50\% accuracy versus CoT’s 85.01\%, alongside a 68\% reduction in tokens. Results on LLaMA-3.1 and 3.2 indicate stronger compression (up to 78\%) but slightly wider accuracy gaps (up to 3\%). These findings confirm that SoT performs reliably across model families, consistently achieving strong token reductions with minimal accuracy degradation.

\subsection{Paradigm-Task Performance}

Task-level results indicate that SoT’s effectiveness is most pronounced in reasoning settings with inherently compressible logic. In mathematical tasks, SoT closely matches the performance of CoT in the majority of settings. For example, in the Qwen-2.5-32B setting, SoT achieves 86.94\% accuracy compared to 84.17\% for CoT, while reducing average output length from 222 to 88 tokens. These gains are attributable to the effectiveness of the \textit{Chunked Symbolism} paradigm in representing arithmetic reasoning concisely, which is the dominant paradigm for this category of reasoning (see Appendix~\ref{sub:paradigm_distribution}).

In commonsense and multi-hop reasoning, SoT maintains strong performance while achieving substantial compression. In the Qwen-2.5-32B setting, SoT reaches 92.00\% accuracy on commonsense tasks using just 34 tokens on average, compared to 91.48\% at 177 tokens under CoT. These improvements are driven by the \textit{Conceptual Chaining} paradigm, which is the prevailing strategy for these reasoning categories and effectively captures structured relationships between ideas.

Domain-specialized tasks, such as PubMedQA and QASC, show more variability in accuracy across models, reflecting the inherent complexity of technical reasoning. Nevertheless, the \textit{Expert Lexicons} paradigm remains effective at compressing domain-specific reasoning, often using half as many tokens as CoT while preserving competitive accuracy. Across all categories, SoT maintains competitive performance with far shorter outputs than CoT, underscoring its adaptive nature.

Further discussion on paradigm distribution across datasets can be found in Appendix~\ref{sec:paradigm_assignments}.

\input{_tables/table_extended_methods}

\subsection{Token-Constrained Alternatives}

Compared to other compression-focused prompting strategies such as Chain-of-Draft (CoD) and Constrained CoT (CCoT), SoT provides a more favorable trade-off between brevity and performance. Although CoD yields the most aggressive reductions in output length, it suffers notable accuracy degradation—for example, a 6.2\% decline on GPT-4o despite a 75\% token reduction. CCoT offers more balanced results, but still lags behind SoT in both efficiency and generalization across reasoning types. Although cases exist where these methods perform better in either accuracy or token reduction, there is no such case where these methods outperform SoT in both. In all observed settings, SoT achieves similar or better accuracy than these methods alongside competitive token reduction.

\subsection{Ensemble Reasoning Methods}
\label{sub:extended_approaches}

To examine SoT’s compatibility with ensemble-style reasoning methods, we integrate it into three established frameworks. Self-Consistency~\cite{self-consistency} aggregates multiple reasoning paths by majority voting to improve answer stability. Self-Refine~\cite{self-refine} enables iterative refinement of reasoning traces through reflection-based prompting. Multi-Agent Debate \cite{multi-agent} simulates deliberation among independent agents, each producing a rationale before converging on a final answer. In each case, we follow the original methodology but substitute SoT in place of CoT as the core reasoning strategy. Further implementation details, including prompts and hyperparameters, are provided in Appendix~\ref{sec:appendix_results}.

Table~\ref{tab:extended_approaches} reports results from integrating SoT into three ensemble reasoning frameworks. In all cases, SoT improves performance relative to CoT, while substantially reducing output length. For instance, in the Self-Refine setting, SoT improves accuracy by 0.27\% while generating 60\% fewer tokens per response. In the Multi-Agent Debate framework, SoT yields a 0.57\% accuracy increase alongside a 69\% token reduction. These results indicate that SoT can be effectively substituted into more complex, multi-pass prompting pipelines, retaining its advantages in both efficiency and output quality.

\input{_tables/table_multilingual_results}

\subsection{Multilingual Reasoning}

To evaluate SoT’s performance in non-English settings, we conduct a multilingual evaluation using Korean, Italian, and German subsets of MMMLU \cite{ds_mmlu}. For each language, we select the same set of 500 questions from each language and generate three responses, for an effective sample size of 1,500. To maintain consistent paradigm selection across languages, each non-English query is paired with its English counterpart and routed using the same routing model. The selected paradigm prompt and associated exemplars are then translated into the target language using GPT-4o~\cite{openai2024gpt4ocard}, preserving both semantic fidelity and structural constraints.

As summarized in Table~\ref{tab:language_results}, SoT reduces output length by over 80\% in all three languages while incurring a modest decrease in accuracy, ranging from -0.33\% to -1.33\%. These findings suggest that the sketching paradigms underlying SoT generalize across linguistic structures and preserve core reasoning logic beyond English.

\input{_tables/table_multimodal_results}

\subsection{Multimodal Reasoning}

To assess SoT’s extensibility to multimodal scenarios, we evaluate its performance using Qwen-2.5-VL-7B~\cite{qwen2.5-VL} on 500 multiple-choice samples from both GQA~\cite{ds_gqa} and the image-based subset of ScienceQA~\cite{ds_scienceqa}. Each sample is run three times for an effective sample size of 1,500. As in the unimodal setting, paradigm selection is handled by the router model. Images and supplementary materials are replaced with a placeholder token during routing (see Section~\ref{sec:paradigm_selection}), allowing the router to focus on the question text. We reuse the same text-only exemplars from the primary experiments.

Results from multimodal evaluations are shown in Table~\ref{tab:multimodal_results}. On ScienceQA, SoT reduces output length by 80\% while outperforming CoT by 6.60\%. On GQA, however, we observed a 2.50\% reduction in accuracy when using SoT while reducing output length by 75\%. The accuracy degradation in GQA likely reflects the difficulty of applying abstract sketching methods to tasks requiring fine-grained visual grounding. Another possible explanation is that the text-only exemplars, while effective in general, may not sufficiently prime the model for vision-intensive reasoning.

\begin{figure}[t]
    \centering
    \includegraphics[width=0.975\linewidth]{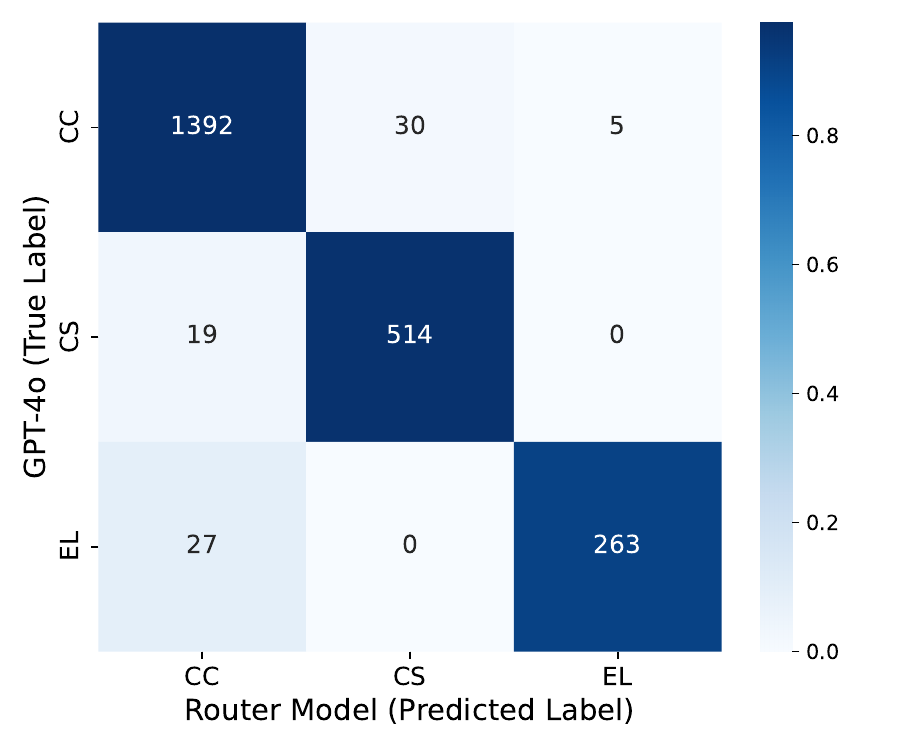}
    \vspace{-0.1in}
    \caption{Confusion matrix illustrating the performance of the router model in selecting among the three SoT paradigms. Predictions are compared against GPT-4o-assigned ground truth labels.}
    \label{fig:router_confusion_appendix}
    \vspace{-0.05in}
\end{figure}

\subsection{Analysis on Routing}
\label{sub:routing_analysis}

To investigate the efficacy of our router model for paradigm assignment, we evaluate its ability to select appropriate reasoning paradigms across the 2,250 samples used in our primary experiments (see Section~\ref{sec:datasets}). Ground-truth labels are produced by GPT-4o using the same labeling protocol as during training (see Section~\ref{sec:paradigm_selection}). As shown in Figure~\ref{fig:router_confusion_appendix}, the model achieves 96.4\% overall accuracy, with high recall for the two most common paradigms, \textit{Conceptual Chaining} (0.964) and \textit{Chunked Symbolism} (0.975). Recall for \textit{Expert Lexicons} is slightly lower at 0.907, largely due to class imbalance. However, this asymmetry is expected as \textit{Expert Lexicons} is intentionally applied more conservatively given its specialized nature, and the router defaults to general paradigms in ambiguous cases to reduce risk of misapplication.

\subsection{Paradigm-Task Alignment}
\label{sub:paradigm_task_alignment}

To test if there is a significant difference between the performance of each paradigm in their respective tasks, we benchmark the performance of all three paradigms on datasets across different reasoning tasks. For any given dataset, we define the dominant paradigm as the paradigm which is assigned to the majority of samples. For example, from the paradigm definitions outlined in Section~\ref{sec:paradigms}, we can assume that the expected-dominant paradigm of GSM8K is \textit{Chunked Symbolism}. In Appendix~\ref{sub:paradigm_distribution}, we conduct an analysis of the expected versus actual dominant paradigms across all datasets to validate the router's overall performance. In all cases the expected-dominant paradigm aligns with the actual-dominant paradigm.

However, this analysis of paradigm routing says nothing of the accuracy on these tasks. For this, we select one representative dataset per paradigm using the previously-mentioned dominant-paradigm distribution in Table~\ref{tab:paradigm_alignment_reasoning_type}. We then run inference on these datasets using all paradigms and compare their performance in Table~\ref{tab:paradigm_performance_dominant}. Our findings indicate that, for all examined datasets, the expected-dominant paradigm outperforms all others in terms of task accuracy. Notably, a paradigm being dominant does not mean it will have the lowest token usage. For example, in medical reasoning, \textit{Expert Lexicons} has the highest accuracy with 85.70\% and, while \textit{Chunked Symbolism} has the lowest token usage, its accuracy is far lower at 73.10\%. These results demonstrate that different reasoning paradigms yield different performance levels depending on the task, and that selecting the optimal paradigm is critical for maximizing accuracy.

\input{_tables/table_paradigm_specific_performance}

%% file: _tables/table_extended_methods.tex
\begin{table}[t]
\centering
\small
\caption{\textbf{Results of Ensemble Methods.} Comparison of SoT and CoT in ensemble reasoning pipelines.}
\resizebox{0.475\textwidth}{!}{
\begin{tabular}{lccccc}
\toprule
\textbf{Approach} & \textbf{Method} & \textbf{Tkn} & \textbf{Acc} & \textbf{Red.} & \textbf{$\Delta$} \\
\midrule
\multirow{2}{*}{\makecell[l]{Self-\\Consistency}} & CoT & 680 & 81.86 & -- & -- \\
 & SoT & 176 & 81.90 & \textbf{74.1} & \textbf{0.04} \\
\midrule
\multirow{2}{*}{\makecell[l]{Self-\\Refine}} & CoT & 614 & 80.53 & -- & -- \\
 & SoT & 244 & 80.80 & \textbf{60.3} & \textbf{0.27} \\
\midrule
\multirow{2}{*}{\makecell[l]{Multi-Agent\\Debate}} & CoT & 766 & 81.87 & -- & -- \\
 & SoT & 238 & 82.44 & \textbf{68.9} & \textbf{0.57} \\
\bottomrule
\end{tabular}
}
\label{tab:extended_approaches}
\end{table}

%% file: _tables/table_multilingual_results.tex
\begin{table}[t]
\centering
\small
\caption{\textbf{Multilingual Results.} Performance comparison of CoT and SoT across different languages.}
\resizebox{0.475\textwidth}{!}{
\begin{tabular}{lccccc}
\toprule
\textbf{Lang.} & \textbf{Method} & \textbf{Tkn} & \textbf{Acc} & \textbf{Red.} & \textbf{$\Delta$} \\
\midrule
\multirow{2}{*}{Korean} & CoT & 308 & 74.20 & -- & -- \\
 & SoT & 49 & 73.40 & \textbf{84.09} & \textbf{-0.80} \\
\midrule
\multirow{2}{*}{Italian} & CoT & 332 & 76.40 & -- & -- \\
 & SoT & 57 & 75.07 & \textbf{82.83} & \textbf{-1.33} \\
\midrule
\multirow{2}{*}{German} & CoT & 306 & 76.40 & -- & -- \\
 & SoT & 48 & 76.07 & \textbf{84.31} & \textbf{-0.33} \\
\bottomrule
\end{tabular}
}
\label{tab:language_results}
\end{table}

%% file: _tables/table_multimodal_results.tex
\begin{table}[t]
\centering
\small
\caption{\textbf{Multimodal Results.} Performance comparison of CoT and SoT for multimodal reasoning tasks.}
\resizebox{0.475\textwidth}{!}{
\begin{tabular}{lccccc}
\toprule
\textbf{Dataset} & \textbf{Method} & \textbf{Tkn} & \textbf{Acc} & \textbf{Red.} & \textbf{$\Delta$} \\
\midrule
\multirow{2}{*}{\makecell[l]{ScienceQA}} & CoT & 136 & 79.60 & -- & -- \\
& SoT & 27 & 86.20 & \textbf{80.15} & \textbf{6.60} \\
\midrule
\multirow{2}{*}{GQA} & CoT & 79 & 74.47 & -- & -- \\
& SoT & 19 & 71.93 & \textbf{75.95} & \textbf{-2.50} \\
\bottomrule
\end{tabular}
}
\label{tab:multimodal_results}
\end{table}

%% file: _tables/table_paradigm_specific_performance.tex
\begin{table}[t]
\centering
\small
\caption{\textbf{Paradigm-Task Alignment.} Comparison of paradigm performance on various reasoning types. Expected-dominant paradigm is in \textbf{bold}.}
\resizebox{0.475\textwidth}{!}{
\begin{tabular}{llcc}
\toprule
\textbf{Reasoning Type} & \textbf{Paradigm} & \textbf{Tkn} & \textbf{Acc} \\
\midrule
\multirow{3}{*}{\makecell[l]{Mathematical\\\textit{SVAMP}}} & \textbf{Chunked Symbolism} & \textbf{30} & \textbf{93.70} \\
 & Expert Lexicons & 33 & 92.60  \\
 & Conceptual Chaining & 34 & 92.00  \\
\midrule

\multirow{3}{*}{\makecell[l]{Medical\\\textit{MedQA}}} & Chunked Symbolism & 46 & 73.10 \\
 & \textbf{Expert Lexicons} & \textbf{52} & \textbf{85.70} \\
 & Conceptual Chaining & 73 & 81.10  \\
 
\midrule
 \multirow{3}{*}{\makecell[l]{Commonsense\\\textit{CommonsenseQA}}} & Chunked Symbolism & 8 & 40.00 \\
 & Expert Lexicons & 18 & 82.90  \\
 & \textbf{Conceptual Chaining} & \textbf{21} & \textbf{84.60} \\
\bottomrule
\end{tabular}
}
\label{tab:paradigm_performance_dominant}
\end{table}

%% file: sections/05_related_work.tex
\section{Related Work}
\label{sec:related}

\paragraph{Token-Efficient Reasoning}

A growing body of work targets the reduction of output length during language model reasoning. Concise Chain-of-Thought~\cite{renze2024ccot} and Constrained CoT ~\cite{ccot-45} apply fixed constraints on the number of steps or words in the reasoning trace. SCOTT~\cite{wang2023scott} uses a two-stage summarization pipeline that compresses verbose CoT outputs into shorter versions. While these methods reduce token usage, they rely on surface heuristics or summary-based rewriting, often reducing clarity. As an orthogonal direction, Coconut~\cite{hao2024traininglargelanguagemodels} bypasses token-based reasoning by operating entirely in the latent vector space, though this requires additional training procedures, making it inapplicable to frozen LLMs. In contrast, SoT rewrites reasoning steps using compact representations, yielding outputs that are both concise and interpretable.

\paragraph{Structured Reasoning Strategies}

Other approaches enhance reasoning by restructuring the generation process itself. Tree-of-Thoughts~\cite{tot} and Graph-of-Thoughts~\cite{got} treat reasoning as a search over intermediate steps, producing graph-structured outputs. Self-Consistency~\cite{self-consistency} improves stability by sampling multiple reasoning paths and selecting the majority answer. While these methods improve accuracy on certain tasks, they often incur significant increases in compute overhead. In contrast, SoT leverages a standard prompting interface to restructure internal reasoning, achieving efficiency gains without increasing inference complexity.

\paragraph{Prompt Compression and Adaptive Inference}

Several techniques improve efficiency through prompt compression or selective computation. Chain-of-Draft (CoD)~\cite{cod} uses densely packed natural language reasoning to reduce length, but this often comes at the cost of clarity and yields large performance drops on more complex reasoning tasks. CoT-Influx~\cite{huang2024fewer} and LLMLingua~\cite{jiang2023llmlingua} prune or compress input exemplars to reduce prompt length. Cascaded inference~\cite{yue2024cascades} and compute-adaptive methods~\cite{arora2025training} dynamically route examples to high-cost inference pipelines only when necessary. SoT differs by addressing compression as a representational design challenge: instead of relying on pruning or selection, it restructures how reasoning is expressed, guided by task-specific cognitive principles.

%% file: sections/06_conclusion.tex
\section{Conclusion}
\label{sec:conclusion}

We present Sketch-of-Thought (SoT), a prompting framework that reduces token usage in language model reasoning by up to 84\%, preserving accuracy in most tasks and incurring only minor trade-offs in others. SoT leverages cognitively inspired paradigms to generate compact yet semantically faithful reasoning traces, offering a practical alternative to verbose prompting. Extensive experiments across 18 reasoning datasets, multiple languages, and multimodal tasks demonstrate SoT’s broad applicability. Its compatibility with ensemble prompting strategies further reinforces its practical utility, particularly in resource-constrained settings. By reframing efficiency as a reasoning design challenge rather than a surface-level compression problem, SoT opens new directions for scalable, cognitively informed prompting.

%% file: sections/07_limitations_future_work.tex
\section*{Limitations and Future Work}
\label{sec:limitations}

Sketch-of-Thought (SoT) is designed for interpretable, efficient reasoning, and while our current approach performs well on a variety of tasks, there exist several interesting directions for future work.

Following prior work, our use of fixed exemplars per paradigm—intentionally chosen to preserve stylistic consistency and interpretability—may limit adaptability to subtle variations within a task type. Alternatively, a retrieval system could dynamically pull in-context exemplars from a larger pool based on the reasoning paradigm and question characteristics. These strategies could help to improve SoT's flexibility across subtly different queries but also disparate tasks and domains.

Also, while we focused this work on evaluating cognitively grounded, prompt-based three paradigms, the framework is not limited to the three we present here. Future work may incorporate additional reasoning paradigms to better adapt SoT to downstream tasks such as code generation. These can be integrated by adding new paradigms, updating the sketching pool, and retraining the routing module accordingly.

Lastly, while our current multilingual experiments already demonstrate SoT’s stability across widely-spoken languages, evaluating its impact in low-resource languages is an exciting direction for future work.

%% file: sections/08_ethics_statement.tex
\section*{Ethics Statement}

This work builds on widely used public datasets and large language models (LLMs). All datasets used in our experiments are publicly available and cited accordingly. Where applicable, we follow dataset authors' intended uses and licensing terms. All models are used in accordance with their respective licenses.

While Sketch-of-Thought (SoT) improves the efficiency of model reasoning, we acknowledge that compressing intermediate outputs may affect interpretability in certain high-stakes settings. We encourage caution when applying SoT in domains such as healthcare or legal analysis, where full transparency of reasoning steps may be essential.

Further, our router model was trained using annotations generated via GPT-4o, and as such may reflect biases present in the underlying model. We recommend further evaluation before deploying SoT in sensitive or high-stakes settings.

%% file: sections/09_acknowledgments.tex
\section*{Acknowledgments}

This work was supported by the Institute for Information \& Communications Technology Planning \& Evaluation (IITP) grant funded by the Korea government (MSIT) (RS-2019-II190075, Artificial Intelligence Graduate School Program (KAIST), and RS-2022-II220713, Meta-learning Applicable to Real-world Problems); the National Research Foundation of Korea (NRF) grant funded by the Korea government (MSIT) (RS-2023-00256259); the grant of the Korea Machine Learning Ledger Orchestration for Drug Discovery Project (K-MELLODDY) funded by the Ministry of Health \& Welfare and the Ministry of Science and ICT, Republic of Korea (RS-2024-00460870); Institute of Information \& Communications Technology Planning \& Evaluation (IITP) with the grant funded by the Ministry of Science and ICT (MSIT) of the Republic of Korea in connection with the Global AI Frontier Lab International Collaborative Research (RS-2024-00469482 \& RS-2024-00509279); and the Artificial Intelligence Industrial Convergence Cluster Development Project funded by the Ministry of Science and ICT (MSIT, Korea) \& Gwangju Metropolitan City.

%% file: appendices/APP_01_AdditionalInformation.tex
\section{Experimental Setup}
\label{sec:appendix_experimental_setup}

\let\thefootnote\relax\footnotetext{GitHub: \url{github.com/SimonAytes/SoT}}
\let\thefootnote\relax\footnotetext{Router: \url{huggingface.co/saytes/SoT_DistilBERT}}

\subsection{Datasets}
\label{ap_datasets}

All datasets used in our experiments are publicly available and accessed via Hugging Face using the dataset IDs listed in Table~\ref{tab:dataset_information_combined}. Where datasets included multiple subsets, we explicitly specified which subset was used in our experiments. All datasets are used in accordance with their respective licenses and terms of use.

\subsection{Model Checkpoints}
\label{ap_models}

We use the following model checkpoints in our experiments:

\noindent\textbf{Qwen 2.5 Family}
\vspace{-0.1in}
\begin{itemize}
    \setlength\itemsep{0.5em}
    \item \texttt{Qwen/Qwen2.5-7B-Instruct}
    \item \texttt{Qwen/Qwen2.5-14B-Instruct}
    \item \texttt{Qwen/Qwen2.5-32B-Instruct}
    \item \texttt{Qwen/Qwen2.5-VL-7B-Instruct}
\end{itemize}

\noindent\textbf{Llama 3 Family}
\vspace{-0.1in}
\begin{itemize}
    \setlength\itemsep{0.5em}
    \item \texttt{meta-llama/Meta-Llama-3-8B-Instruct}
    \item \texttt{meta-llama/Meta-Llama-3-11B-Instruct}
\end{itemize}

\noindent\textbf{Closed-source Models}
\vspace{-0.1in}
\begin{itemize}
    \setlength\itemsep{0.5em}    
    \item \texttt{gpt-4o-2024-11-20}
    \item \texttt{claude-3-5-sonnet-20241022}
    \vspace{-0.05in}
\end{itemize}

All open-weight models were accessed through Hugging Face via the transformers library \cite{transformers_lib} and evaluated in their instruction-tuned form. Closed-weight models such as GPT-4o and Claude Sonnet 3.5 were accessed through their respective Python API wrappers. All models are used in accordance with their licenses.

\subsection{Inference Environment}
\label{ap_inference}

All experiments were conducted on 2 x A5000 24GB GPUs on a Linux distribution running CUDA 12.1. For inference, we use FlashAttention2 \cite{flashattention2} for acceleration. All models were run in bfloat16 precision where supported. No parameter fine-tuning or additional adaptation was applied to the LLMs during experimentation.

\subsection{Reproducibility}
\label{ap_reproducibility}

All experiments were conducted with the same fixed random seed, $42$, to ensure reproducibility across runs. We used a consistent temperature of 0.5 for all models across all methods and tasks. For few-shot setups, exemplars were selected prior to evaluation and held constant across all trials. Token counts were measured using the default tokenizer associated with each model’s checkpoint. For closed source models accessed via the API, token counts were obtained through the token logs found in the returned inference object.

\input{_tables/APPENDIX_Datasets}

%% file: _tables/APPENDIX_Datasets.tex
\begin{table*}[t]
\centering
\scriptsize
\renewcommand{\arraystretch}{0.8}
\setlength{\tabcolsep}{3.0pt}
\caption{\textbf{Dataset Information.} Comprehensive details of datasets used for our experiments.}
\vspace{-0.05in}
\resizebox{\textwidth}{!}{
\begin{tabular}{@{}lcccccc@{}}
\toprule
\textbf{Dataset} & \textbf{Citation} & \textbf{HF ID} & \textbf{Train Split:Subset} & \textbf{Train Size} & \textbf{Test Split:Subset} & \textbf{Test Size} \\
\midrule
GSM8K & \citet{ds_gsm8k} & \texttt{gsm8k} & \texttt{main:train} & 1000 & \texttt{main:test} & 150 \\
SVAMP & \citet{ds_svamp} & \texttt{ChilleD/SVAMP} & \texttt{train} & 700 & \texttt{test} & 150 \\
AQUA-RAT & \citet{ds_aqua_rat} & \texttt{aqua\_rat} & \texttt{train} & 1000 & \texttt{test} & 150 \\
DROP & \citet{ds_drop} & \texttt{drop} & \texttt{train} & 1000 & \texttt{validation} & 150 \\
OpenbookQA & \citet{ds_openbookqa} & \texttt{openbookqa} & \texttt{train} & 1000 & \texttt{test} & 150 \\
StrategyQA & \citet{ds_strategyqa} & \texttt{ChilleD/StrategyQA} & \texttt{train} & 1000 & \texttt{test} & 150 \\
LogiQA & \citet{ds_logiqa} & \texttt{lucasmccabe/logiqa} & \texttt{train} & 1000 & \texttt{test} & 150 \\
Reclor & \citet{ds_reclor} & \texttt{metaeval/reclor} & \texttt{train} & 1000 & \texttt{validation} & 150 \\
HotPotQA & \citet{ds_hotpotqa} & \texttt{hotpot\_qa} & \texttt{distractor:train} & 1000 & \texttt{distractor:validation} & 150 \\
MuSiQue-Ans & \citet{ds_musique_ans} & \texttt{dgslibisey/MuSiQue} & \texttt{train} & 1000 & \texttt{validation} & 150 \\
QASC & \citet{ds_qasc} & \texttt{allenai/qasc} & \texttt{train} & 1000 & \texttt{validation} & 150 \\
Worldtree & \citet{ds_worldtree} & \texttt{nguyen-brat/worldtree} & \texttt{train (last 1000)} & 1000 & \texttt{train (rest)} & 150 \\
PubMedQA & \citet{ds_pubmedqa} & \texttt{qiaojin/PubMedQA} & \texttt{pqa\_labeled (last 500)} & 500 & \texttt{pqa\_labeled (first 150)} & 150 \\
MedQA & \citet{ds_medqa} & \texttt{bigbio/med\_qa} & \texttt{med\_qa\_en\_source:train} & 1000 & \texttt{med\_qa\_en\_source:validation} & 150 \\
CommonsenseQA & \citet{ds_commonsenseqa} & \texttt{tau/commonsense\_qa} & \texttt{train} & 1000 & \texttt{validation} & 150 \\
MMLU & \citet{ds_mmlu} & \texttt{cais/mmlu} & — & — & \texttt{test:all} & 500 \\
MMMLU & \citet{ds_mmlu} & \texttt{openai/MMMLU} & — & — & \texttt{test:KO\_KR, DE\_DE, IT\_IT} & 500 \\
ScienceQA & \citet{ds_scienceqa} & \texttt{lmms-lab/ScienceQA} & — & — & \texttt{test:ScienceQA-IMG} & 500 \\
GQA & \citet{ds_gqa} & \texttt{lmms-lab/GQA} & — & — & \texttt{val:val\_all\_images} & 500 \\
\bottomrule
\end{tabular}
}
\vspace{-0.05in}
\label{tab:dataset_information_combined}
\end{table*}

%% file: appendices/APP_02_SystemPromptsExemplars.tex
\section{Prompting Framework}
\label{sec:appendix_prompts}

\subsection{System Prompt Format}
\label{sub:prompt_format}

Each paradigm-specific system prompt follows a consistent structure composed of four sections. Among these sections, \textit{Role \& Objective} is the most extensive. This design choice is motivated by two factors. First, since our method is training-free and prompt-based, we found that smaller LLMs require more explicit and structured instructions to reliably follow the desired output format. In early trials, such models often ignored or deviated from intended behavior when given minimal guidance. Second, recent work on structural priming in LLMs shows that lexical and semantic cues in preceding context influence syntactic expectations and downstream predictions \citep{semanticpriming2024}. We leverage this by “setting the stage” for the model to reason within the appropriate paradigm.

\paragraph{Role \& Objective}  
Provides background on the paradigm, including its cognitive basis and theoretical motivation. It outlines representative use cases and serves as a semantic primer to help align the model's reasoning style with the paradigm.

\paragraph{Application Steps}  
Describes a step-by-step procedure for applying the paradigm to solve a problem. This includes identifying relevant concepts, performing transformations, and following best practices for structuring the reasoning process.

\paragraph{Rules \& Directives}  
Specifies required tone, structure, and formatting constraints. It highlights common failure modes—such as verbosity, redundancy, or incorrect notation—and explicitly defines output style requirements (see Appendix~\ref{sub:output_format}).

\paragraph{Closing Statement}  
Ends with a reminder to adhere to the formatting guidelines, reinforcing the objective of concise, structured reasoning.

\subsection{Output Conventions}
\label{sub:output_format}

To ensure consistent evaluation and accurate token-level comparisons, all outputs follow a strict formatting protocol:
\vspace{-0.05in}
\begin{itemize}
    \setlength\itemsep{0.25em}
    \item \textbf{Answers} must be enclosed in \texttt{\textbackslash boxed\{\ldots\}}.
    \item \textbf{Reasoning traces} must appear within \texttt{<think>} and \texttt{</think>} tags.
\end{itemize}

This formatting allows for reliable programmatic parsing and segmentation of outputs into intermediate reasoning and final answers, supporting reproducibility and enabling fair evaluation across prompting methods and models.

Because all experiments are conducted using instruction-tuned LLMs with no additional fine-tuning to enforce output structure, we explicitly reserve space in both the \textit{Rules \& Directives} and \textit{Closing Statement} sections of each system prompt to reinforce these formatting requirements.

In practice, we find that providing exemplars alone is insufficient for enforcing consistent formatting. In early experiments, models frequently omitted structural tags or deviated from the expected format when prompted using exemplars only. After incorporating explicit formatting instructions into the system prompt, the rate of malformed or non-compliant outputs dropped to near zero across all paradigms and model variants.

\subsection{Paradigm Prompts}
\label{sub:paradigm_prompts}

We provide reference versions of our paradigm system prompts for \textit{Conceptual Chaining}, \textit{Chunked Symbolism}, and \textit{Expert Lexicons} in Figures \ref{fig:conceputal_chaining_prompts}, \ref{fig:chunked_symbolism_prompts}, and \ref{fig:expert_lexicons_prompts}, respectively. Parts of the prompts have been adjusted to render correctly in this document. We direct the interested reader to our public code repositories for full, code-ready prompts.

\subsection{In-Context Exemplars}
\label{sub:paradigm_exemplars}

We provide three in-context exemplars for each method evaluated in our study to guide model outputs during inference. For Sketch-of-Thought (SoT), a separate set of exemplars is constructed for each paradigm to match the distinct reasoning styles each paradigm is designed to elicit. Example questions are manually selected to reflect typical tasks associated with each paradigm's target use cases. To construct exemplars, we first generate candidate responses using Qwen-2.5-32B with the corresponding system prompt, then manually select outputs that most faithfully demonstrate the paradigm's intended structure, clarity, and conciseness. This results in three curated exemplars per paradigm.

For baseline methods—Chain-of-Thought (CoT), Constrained CoT (CCoT), and Chain-of-Draft (CoD)—we apply the same method. Each method is prompted using its respective strategy, and the most stylistically representative outputs are selected. Because these baselines do not dynamically adapt to the query type, we ensure fair coverage by drawing exemplars from the same three reasoning categories used for SoT (e.g., commonsense, mathematical, medical). One exemplar is selected per category, yielding a total of three per method. All exemplars are held fixed across all experiments.

\subsection{Combined System Prompt}
\label{sub:combined_prompt}

The question naturally arises as to whether the Sketch-of-Thought (SoT) framework would have benefited from a more streamlined approach in which the LLM is prompted with a single joint prompt describing all three paradigms, allowing the model to select one or apply them jointly. While promising in principle, our exploration of this approach in early development resulted in high rates of formatting errors, especially for models with fewer than 70B parameters. These smaller models frequently failed to follow paradigm-specific instructions or ignored the structured reasoning techniques altogether, reverting to verbose or default reasoning styles. We believe this reflects a limitation of training-free prompting for smaller models. Furthermore, as discussed in our future work section, SoT is designed to scale to many more than three paradigms. In that setting, a joint prompt would incur significant token overhead from combining system prompts and exemplars, undercutting the efficiency gains SoT is designed to provide.

\subsection{Classification Prompt}
\label{sub:classification_prompt}

The router model used to assign paradigms was trained using GPT-4o-generated labels. The classification prompt presented each query and instructed the model to assign one of the three paradigms based on reasoning characteristics, following the heuristic definitions given in Section~\ref{sec:paradigms}. A reference version of the classification prompt is shown in Figure~\ref{fig:classification_prompt}. To conserve space we omit repetitive text in this version. We direct the interested reader to our public code repositories for the full, unabridged classification prompt.

%% file: appendices/APP_04_ParadigmAssignments.tex
\section{Paradigm Assignment Analysis}
\label{sec:paradigm_assignments}

\subsection{Router Architecture Ablation}
\label{sub:router_architecture_ablation}

\input{_tables/APPENDIX_Router_Model_Selection}

To support our choice of architecture, we conducted an ablation comparing four base models for routing: DistilBERT ~\citep{distilbert}, GPT-2 ~\citep{gpt-2}, BERT-base, and BERT-large ~\citep{bert-models}. All models were trained on the same task with identical supervision and hyperparameters. We measured test accuracy, latency, and peak memory usage during inference for the classification of 1,250 samples from our test set. The results of this ablation are shown in Table~\ref{tab:router_model_selection}.

While GPT-2 and BERT-base achieved slightly higher accuracy than DistilBERT, they also incurred significantly higher memory demand and higher latency. BERT-large had the highest parameter count but the lowest accuracy. Overall, DistilBERT achieved the best tradeoff across accuracy, latency, and memory overhead, confirming that it is the ideal architecture for the purpose of our experiments.

\subsection{Analysis of Machine Labeling}
\label{sub:gpt4o_human_evaluation}

As outlined in Section~\ref{sec:paradigm_selection} and~\ref{sub:routing_analysis}, our method relies on using GPT-4o~\citep{openai2024gpt4ocard} for machine-labeling tasks—both for the training and evaluation of our router. To evaluate the quality of GPT-4o’s paradigm assignment, we designed a human study to measure agreement between human annotators and GPT-4o. We randomly selected 30 samples—uniformly distributed across the three paradigms—and recruited nine annotators with no prior hands-on knowledge of the Sketch-of-Thought framework. Annotators were given the exact same system prompt used by GPT-4o and were allowed to reference it freely while annotating (see Appendix~\ref{sub:classification_prompt}).

We evaluated the agreement between the human annotators and machine labeling with Cohen's Kappa ($k$). The results from this experiment show substantial agreement between GPT-4o and the human-majority label ($k=0.95$), with strong inter-human agreement ($k=0.72$) supporting the reliability of our human study and highlighting the quality of the machine-labeled training data.

\subsection{Routing Distribution by Dataset}
\label{sub:paradigm_distribution}

To better understand how SoT paradigms align with reasoning task types, we analyze the output of our router model across the datasets used in our primary experiments (Section~\ref{sec:datasets}). Table~\ref{tab:paradigm_alignment_reasoning_type} reports the predicted paradigm distribution, dominant paradigm, and its agreement with an expected paradigm label defined based on the paradigm descriptions in Section~\ref{sec:paradigms}. These counts reflect router predictions over the 150 samples used per dataset in our primary experiments. The dominant paradigm is defined as the one with the highest frequency within a dataset, and we compare this to the expected paradigm, which is assigned based on prior task categorizations and paradigm design goals.

The router’s predictions match expectations in all 15 datasets, with 100\% agreement between dominant and expected paradigms. Most datasets are routed to a single paradigm, reflecting high confidence and class purity. In a few edge cases (e.g., DROP, LogiQA, QASC), we observe minor cross-paradigm overlap, though these do not shift the dominant label. This behavior aligns with our router’s conservative design, which favors general-purpose paradigms (especially \textit{Conceptual Chaining}) in ambiguous scenarios.

As expected, \textit{Conceptual Chaining} dominates in commonsense, logical, and multi-hop datasets (e.g., StrategyQA, HotPotQA, Reclor), where relational inference is critical. 

\textit{Chunked Symbolism} is used exclusively in mathematical tasks (e.g., GSM8K, AQUA, SVAMP), where symbolic notation offers the clearest compression benefit. In DROP, which mixes symbolic and textual reasoning, some samples are routed to \textit{Conceptual Chaining}, reflecting the complementary reasoning patterns among the paradigms. 

\textit{Expert Lexicons} is most common in domain-specific datasets like PubMedQA and MedQA. Occasional routing to \textit{Conceptual Chaining} in these cases reflects the router’s conservative fallback behavior, favoring general-purpose paradigms when confidence is low—a design choice that reduces the risk of applying technical conventions in inappropriate contexts.

\subsection{Paradigm Assignment Examples}
\label{sub:paradigm_case_examples}

To illustrate how the SoT router assigns paradigms to diverse questions, we present three representative examples below—one from each SoT paradigm. Each example includes the query as processed by the router and the assigned paradigm. Appendix~\ref{sub:misalignment_example} showcases an edge case that demonstrates the system's conservative fallback behavior.

\begin{center}
\begin{promptbox}[Chunked Symbolism (GSM8K)]
\textbf{Query:}\newline
Darrell and Allen's ages are in the ratio of 7:11. If their total age now is 162, calculate Allen's age 10 years from now.\newline

\textbf{Assigned Paradigm:}\newline
Chunked Symbolism
\end{promptbox}
\end{center}

\begin{center}
\begin{promptbox}[Conceptual Chaining (OpenbookQA)]
\textbf{Query:}\newline
Polar bears require\newline
Choices:\newline
A. a tropical environment\newline
B. a frigid environment\newline
C. a tepid environment\newline
D. a warm environment\newline

\textbf{Assigned Paradigm:}\newline
Conceptual Chaining
\end{promptbox}
\end{center}

\begin{center}
\begin{promptbox}[Expert Lexicons (PubMedQA)]
\textbf{Query:}\newline
Is the holmium:YAG laser the best intracorporeal lithotripter for the ureter?\newline
Choices: Yes, No, Maybe\newline

\textbf{Assigned Paradigm:}\newline
Expert Lexicons
\end{promptbox}
\end{center}

\subsection{Router Misalignment}
\label{sub:misalignment_example}

The results from Section~\ref{sub:paradigm_task_alignment} show that using the correct paradigm consistently yields the highest accuracy. While the drop in performance varies by task, the trend is consistent across datasets. Notably, output token usage remained relatively stable across paradigms due to fixed prompts and exemplar lengths, suggesting that routing errors primarily affect accuracy rather than efficiency.

To examine a case in which the predicted paradigm of a query differs from its parent dataset's dominant paradigm, we consider the following.

\begin{center}
\begin{promptbox}[Edge Case: Conceptual Chaining (PubMedQA)]
\textbf{Query:}\newline
[Context Here] Question: Birth characteristics and risk of low intellectual performance in early adulthood: are the associations confounded by socioeconomic factors in adolescence or familial effects?\newline
Choices: Yes, No, Maybe\newline

\textbf{Assigned Paradigm:}\newline
Conceptual Chaining
\end{promptbox}
\end{center}

For this query, the router assigns the query to \textit{Conceptual Chaining} rather than \textit{Expert Lexicons}, despite the latter being the dominant paradigm for that dataset (see Appendix~\ref{sub:paradigm_distribution}). However, this is not necessarily an incorrect prediction. The question centers on causal inference, which aligns with \textit{Conceptual Chaining}. Although the context is medical, the query does not rely on the domain-specific jargon typical of \textit{Expert Lexicons}. Because PubMedQA blends domain expertise with general causal reasoning, it contains a nontrivial number of such mixed-paradigm cases (approximately 66\% \textit{Expert Lexicons}, 33\% \textit{Conceptual Chaining}).

%% file: _tables/APPENDIX_Router_Model_Selection.tex
\begin{table}[t]
\centering
\small
\caption{\textbf{Router Ablation Results.} Comparison of benchmark performance among candidate router model architectures. Accuracy and latency values are averaged across all test cases. VRAM denotes the peak memory usage of the model during the experiment in megabytes.}
\resizebox{0.475\textwidth}{!}{
\begin{tabular}{lccccc}
\toprule
\textbf{Model} & \textbf{Params} & \textbf{Acc} & \textbf{Latency (s)} & \textbf{VRAM} \\
\midrule
DistilBERT & 67M & 90.31 & 0.0118 & 283 MB \\
GPT-2 & 137M & 91.11 & 0.0107 & 652 MB \\
BERT-base & 110M & 90.93 & 0.0139 & 445 MB \\
BERT-large & 336M & 88.93 & 0.0259 & 1309 MB \\
\bottomrule
\end{tabular}
}
\vspace{-0.05in}
\label{tab:router_model_selection}
\end{table}

%% file: appendices/APP_03_AdditionalExperiments.tex
\section{Extended Results}
\label{sec:appendix_results}

\subsection{Per-Dataset Results}
\label{sub:primary_dataset_results}

We report per-dataset results from the primary experiments across all model families and prompting strategies in Tables~\ref{tab:full_results_qwen} and~\ref{tab:full_results_llama_closed}. Results are shown for each individual dataset with accuracy and token counts averaged across three runs per dataset. For further information regarding the primary experiments, see Section~\ref{sec:experiments} for the experimental design and Section~\ref{sec:results} for the results and discussion.

\subsection{Multi-Agent Debate}
\label{sub:multi_agent_debate}

To evaluate whether Sketch-of-Thought (SoT) remains effective in ensemble-style deliberation, we incorporate it into the Multi-Agent Debate (MAD) framework \cite{multi-agent}. This method simulates independent agents answering the same question and iteratively revising their answers through multi-round critique. For the Multi-Agent Debate setup, we preserve the paradigm-specific SoT system prompt and introduce a debate prompt that allows agents to revise their reasoning in response to other agents' answers. The debate prompt is structured to request updated responses while retaining the specified output formatting conventions.

\begin{center}
\begin{promptbox}[Multi-Agent Debate Prompt]
You are participating in a multi-agent debate. Other agents have responded as follows:

\#Agent 1:\newline
<think>
[agent 1's reasoning]
</think>\newline
Answer: [agent 1's answer]

Your previous answer was:\newline
<think>
[your previous reasoning]
</think>\newline
Answer: [your previous answer]\newline
Would you like to revise your reasoning or stick with it? Please provide your updated reasoning inside <think>...</think> tags and your final answer inside \texttt{\textbackslash boxed\{\}}.
\end{promptbox}
\end{center}

Each debate run involves three agents and a maximum of three rounds. In the first round, all agents independently generate answers using SoT prompts selected by the router model (see Section~\ref{sec:paradigm_selection}). In subsequent rounds, each agent receives the other agents’ reasoning and has the opportunity to revise its answer using the shared debate prompt above. Debates terminate early if all agents converge on the same answer. If consensus is not reached within three rounds, a majority vote determines the final answer. The rationale of the majority-aligned agent is retained as the final justification.

Notably, we find that using SoT does not have a notable impact on the number of rounds-per-query. For CoT we observed an average of 1.14 rounds-per-query, similar to SoT's 1.11. Results are shown in Table~\ref{tab:extended_approaches} and discussed in Section~\ref{sec:results}.

\subsection{Self-Consistency}
\label{sub:self_consistency}

The Self-Consistency framework \cite{self-consistency} is an ensemble reasoning method where multiple reasoning paths are generated from the same input query and the final answer is chosen via majority vote. This method improves answer stability and can, in some cases, mitigate the randomness that arises from the LLM's sampling parameters (i.e., temperature). For our experiments, we generate three outputs for each sample, extract their answers, and finally select the most-frequent answer. In cases where all three answers are different, we implement a random selection fallback and select the final answer at random. The results for Self-Consistency are reported in Table~\ref{tab:extended_approaches}.

\subsection{Self-Refine}
\label{sub:self_refine}

The Self-Refine framework \cite{self-refine} is a reflection-based prompting strategy in which a single agent critiques and revises its own reasoning trace. Each trial consists of a two-step loop: (1) a critique prompt is applied to the model’s initial response to identify any flaws or ambiguities, and (2) a refinement prompt is used to generate a revised answer based on the critique. The initial reasoning trace is produced using the appropriate SoT paradigm (selected via the router), after which the model reflects on its output and revises it. Prompt details for both critique and refinement phases are provided below, and results are reported in Table~\ref{tab:extended_approaches}.

\begin{center}
\begin{promptbox}[Critique Prompt]
You are reviewing a response generated using the \texttt{<paradigm>} reasoning paradigm for the following question:

Question: \texttt{<question>}

<think>
<original reasoning>
</think>\newline
Answer: <original answer>\newline
Please identify any flaws, gaps, or unclear steps, while maintaining the structured, concise format encouraged by this paradigm. Respond WITHOUT using <think>...</think> tags or \texttt{\textbackslash boxed\{\}}.
\end{promptbox}
\end{center}

\begin{center}
\begin{promptbox}[Refine Prompt]
You are refining a response originally generated using the \texttt{<paradigm>} reasoning paradigm for the following question:

Question: \texttt{<question>}

Original Reasoning:\newline
<think>
<original reasoning>
</think>\newline
Answer: <original answer>\newline
Critique: <model-generated critique>\newline
Please revise the response using the critique provided, ensuring your reasoning remains concise, structured, and consistent with the paradigm. Use <think>...</think> for reasoning and \texttt{\textbackslash boxed\{\}} for the final answer.
\end{promptbox}
\end{center}

%% file: appendices/APP_05_IO_Examples.tex
\section{Output Examples}
\label{sec:appendix_examples}

Figure~\ref{fig:reasoning_trace_examples} presents representative input–output examples for each of the three Sketch-of-Thought (SoT) paradigms alongside outputs from baseline prompting strategies including Chain-of-Thought (CoT), Chain-of-Draft (CoD), and Constrained CoT (CCoT). Compared to baselines, SoT responses are significantly more compact while maintaining logical structure and semantic completeness. While CoD and CCoT reduce length relative to CoT, they rely solely on shortened natural language, often resulting in compressed but less interpretable text.

%% file: _tables/APPENDIX_summary_statistics_QWEN.tex
\clearpage
\begin{table*}[t]
\centering
\scriptsize
\renewcommand{\arraystretch}{0.8}
\setlength{\tabcolsep}{3pt}
\caption{\textbf{Full results for main experiments (Qwen Models).}}
\label{tab:full_results_qwen}
\vspace{-0.05in}
\begin{tabular}{c l l *{12}{c}}
\toprule
\multirow{2}{*}{\makecell{\textbf{Reasoning}\\\textbf{Type}}}
& \multirow{2}{*}{\textbf{Dataset}} 
& \multirow{2}{*}{\textbf{Method}} 
& \multicolumn{4}{c}{\makecell{\textbf{Qwen-7B}}}
& \multicolumn{4}{c}{\makecell{\textbf{Qwen-14B}}} 
& \multicolumn{4}{c}{\makecell{\textbf{Qwen-32B}}} \\
\cmidrule(lr){4-7} \cmidrule(lr){8-11} \cmidrule(lr){12-15}
& & & Acc & $\sigma_{a}$ & Tkn & $\sigma_{t}$ & Acc & $\sigma_{a}$ & Tkn & $\sigma_{t}$ & Acc & $\sigma_{a}$ & Tkn & $\sigma_{t}$ \\
\cmidrule{1-15}

\multirow{16}*{\rotatebox[origin=c]{90}{Mathematical}}
& \multirow{4}*{GSM8K}
& CoT & 87.11 & 2.69 & 211 & 3.77 & 93.56 & 2.04 & 215 & 4.99 & 94.89 & 0.38 & 263 & 3.21 \\
 & & CCoT & 88.67 & 1.33 & 92 & 0.38 & 88.89 & 0.38 & 135 & 0.85 & 83.78 & 2.34 & 86 & 1.19 \\
 & & CoD & 59.56 & 0.38 & 66 & 0.83 & 64.89 & 1.54 & 70 & 1.43 & 67.78 & 2.34 & 58 & 0.59 \\
 & & SoT & 84.22 & 2.69 & 79 & 1.03 & 92.89 & 0.38 & 87 & 0.82 & 95.78 & 0.38 & 103 & 0.98 \\
\cmidrule{2-15}

 & \multirow{4}*{AQUA}
 & CoT & 64.67 & 3.33 & 248 & 12.71 & 78.67 & 0.67 & 267 & 9.08 & 76.89 & 2.52 & 289 & 14.53 \\
 & & CCoT & 68.89 & 0.38 & 111 & 1.05 & 76.22 & 1.02 & 155 & 5.11 & 74.00 & 3.46 & 97 & 1.14 \\
 & & CoD & 61.56 & 1.54 & 72 & 1.35 & 59.33 & 2.91 & 79 & 1.49 & 64.00 & 2.31 & 64 & 0.47 \\
 & & SoT & 70.89 & 4.82 & 106 & 2.99 & 77.56 & 1.02 & 116 & 1.87 & 82.22 & 2.52 & 138 & 2.10 \\
\cmidrule{2-15}

 & \multirow{4}*{SVAMP}
 & CoT & 87.11 & 1.39 & 149 & 2.96 & 92.44 & 1.02 & 136 & 0.91 & 92.22 & 1.68 & 181 & 2.59 \\
 & & CCoT & 84.89 & 1.02 & 61 & 0.61 & 89.33 & 0.67 & 86 & 2.21 & 88.67 & 1.76 & 61 & 0.01 \\
 & & CoD & 80.89 & 2.14 & 43 & 0.55 & 85.78 & 3.15 & 51 & 0.71 & 85.33 & 1.15 & 44 & 0.77 \\
 & & SoT & 90.67 & 0.00 & 49 & 0.72 & 88.89 & 0.77 & 53 & 1.11 & 94.22 & 0.38 & 58 & 1.20 \\
\cmidrule{2-15}

 & \multirow{4}*{DROP}
 & CoT & 72.89 & 2.14 & 136 & 1.81 & 67.33 & 1.76 & 141 & 1.76 & 72.67 & 1.76 & 155 & 2.62 \\
 & & CCoT & 69.56 & 0.38 & 59 & 1.38 & 70.89 & 0.77 & 86 & 1.98 & 75.56 & 1.39 & 60 & 0.74 \\
 & & CoD & 65.33 & 2.40 & 48 & 0.55 & 66.89 & 1.68 & 51 & 1.02 & 70.67 & 1.15 & 47 & 0.33 \\
 & & SoT & 71.33 & 0.67 & 55 & 0.58 & 71.56 & 0.38 & 58 & 0.56 & 75.56 & 0.38 & 55 & 0.56 \\
\midrule

\multirow{12}*{\rotatebox[origin=c]{90}{Commonsense}}
& \multirow{4}*{CommonsenseQA}
& CoT & 84.67 & 4.00 & 176 & 6.17 & 85.78 & 1.92 & 158 & 0.16 & 85.33 & 1.33 & 188 & 1.51 \\
 & & CCoT & 36.22 & 1.68 & 44 & 0.74 & 87.33 & 0.67 & 53 & 1.08 & 82.67 & 1.76 & 48 & 0.11 \\
 & & CoD & 77.56 & 2.04 & 38 & 0.20 & 82.89 & 0.38 & 40 & 1.20 & 83.11 & 0.38 & 38 & 0.43 \\
 & & SoT & 85.78 & 1.02 & 25 & 0.20 & 86.00 & 0.00 & 33 & 0.26 & 86.22 & 1.02 & 29 & 0.94 \\
\cmidrule{2-15}

 & \multirow{4}*{OpenbookQA}
 & CoT & 82.67 & 1.76 & 171 & 6.09 & 95.11 & 1.02 & 154 & 1.58 & 95.33 & 0.67 & 186 & 1.41 \\
 & & CCoT & 68.00 & 3.46 & 44 & 0.23 & 93.78 & 0.38 & 54 & 1.37 & 93.56 & 1.39 & 49 & 0.40 \\
 & & CoD & 88.89 & 0.38 & 38 & 0.22 & 94.22 & 0.77 & 41 & 0.05 & 94.67 & 0.67 & 40 & 0.13 \\
 & & SoT & 85.56 & 0.38 & 28 & 0.56 & 93.56 & 0.38 & 38 & 0.42 & 95.11 & 0.77 & 32 & 0.33 \\
\cmidrule{2-15}

 & \multirow{4}*{StrategyQA}
 & CoT & 92.22 & 0.77 & 128 & 5.44 & 93.33 & 1.33 & 139 & 3.01 & 93.78 & 0.38 & 158 & 1.92 \\
 & & CCoT & 88.22 & 1.92 & 46 & 0.27 & 90.44 & 2.14 & 67 & 0.69 & 90.22 & 1.02 & 51 & 0.72 \\
 & & CoD & 87.78 & 1.92 & 35 & 0.18 & 90.00 & 0.67 & 42 & 0.29 & 90.67 & 1.15 & 38 & 0.10 \\
 & & SoT & 88.89 & 0.38 & 38 & 0.18 & 93.11 & 0.38 & 42 & 0.14 & 94.67 & 0.00 & 40 & 0.14 \\
\midrule

\multirow{8}*{\rotatebox[origin=c]{90}{Logical}}
& \multirow{4}*{LogiQA}
& CoT & 51.78 & 6.74 & 302 & 4.19 & 56.22 & 3.67 & 265 & 3.31 & 60.67 & 1.15 & 306 & 5.75 \\
 & & CCoT & 53.78 & 0.38 & 68 & 1.41 & 60.22 & 1.02 & 104 & 3.90 & 63.11 & 3.42 & 63 & 0.32 \\
 & & CoD & 53.11 & 2.04 & 53 & 0.96 & 54.22 & 0.38 & 52 & 0.01 & 63.11 & 2.04 & 47 & 0.50 \\
 & & SoT & 53.11 & 2.04 & 85 & 2.36 & 56.00 & 1.76 & 75 & 1.13 & 60.22 & 0.38 & 79 & 2.03 \\
\cmidrule{2-15}

 & \multirow{4}*{Reclor}
 & CoT & 79.56 & 3.67 & 266 & 9.36 & 77.78 & 1.68 & 231 & 4.08 & 81.78 & 1.02 & 289 & 1.12 \\
 & & CCoT & 73.56 & 1.68 & 59 & 0.94 & 79.78 & 1.39 & 75 & 1.70 & 82.44 & 0.38 & 57 & 1.03 \\
 & & CoD & 75.56 & 0.77 & 45 & 0.16 & 78.22 & 2.04 & 42 & 0.64 & 82.67 & 2.67 & 43 & 0.41 \\
 & & SoT & 70.89 & 1.39 & 47 & 1.22 & 78.89 & 0.77 & 52 & 1.22 & 81.78 & 1.02 & 53 & 0.46 \\
\midrule

\multirow{8}*{\rotatebox[origin=c]{90}{Multi-Hop}}
& \multirow{4}*{HotPotQA}
& CoT & 89.33 & 1.15 & 124 & 4.33 & 90.00 & 1.15 & 135 & 2.08 & 92.22 & 1.39 & 143 & 1.55 \\
 & & CCoT & 91.56 & 0.77 & 49 & 0.53 & 89.56 & 1.92 & 84 & 0.19 & 93.33 & 0.67 & 51 & 0.20 \\
 & & CoD & 90.22 & 0.38 & 41 & 0.19 & 89.78 & 2.34 & 44 & 0.49 & 91.11 & 0.38 & 39 & 0.09 \\
 & & SoT & 88.22 & 1.68 & 43 & 0.57 & 90.22 & 1.68 & 42 & 0.36 & 94.00 & 0.67 & 41 & 0.06 \\
\cmidrule{2-15}

 & \multirow{4}*{MuSiQue}
 & CoT & 57.11 & 1.68 & 151 & 2.73 & 65.33 & 2.00 & 163 & 1.24 & 66.67 & 0.67 & 167 & 2.95 \\
 & & CCoT & 66.22 & 2.04 & 57 & 0.90 & 68.22 & 1.02 & 98 & 2.87 & 66.89 & 1.92 & 57 & 0.26 \\
 & & CoD & 62.00 & 1.76 & 45 & 0.68 & 71.11 & 2.78 & 47 & 0.58 & 68.89 & 1.02 & 43 & 0.45 \\
 & & SoT & 64.00 & 0.67 & 47 & 0.61 & 69.56 & 0.38 & 49 & 0.37 & 69.78 & 0.38 & 46 & 0.62 \\
\midrule

\multirow{8}*{\rotatebox[origin=c]{90}{Scientific}}
& \multirow{4}*{QASC}
& CoT & 83.78 & 1.02 & 176 & 2.50 & 83.78 & 1.54 & 163 & 1.38 & 87.33 & 1.15 & 222 & 3.67 \\
 & & CCoT & 29.11 & 2.14 & 44 & 0.50 & 82.89 & 2.14 & 54 & 1.23 & 79.11 & 0.38 & 50 & 0.52 \\
 & & CoD & 76.89 & 1.68 & 38 & 0.29 & 82.00 & 1.33 & 45 & 1.62 & 81.33 & 0.00 & 43 & 0.61 \\
 & & SoT & 77.78 & 1.02 & 25 & 0.17 & 82.89 & 1.02 & 36 & 0.09 & 84.22 & 1.02 & 30 & 0.90 \\
\cmidrule{2-15}

 & \multirow{4}*{Worldtree}
 & CoT & 93.56 & 1.68 & 185 & 3.77 & 98.00 & 0.67 & 166 & 1.03 & 98.44 & 0.38 & 204 & 1.41 \\
 & & CCoT & 96.44 & 1.02 & 44 & 0.78 & 96.00 & 1.15 & 56 & 0.46 & 98.67 & 0.00 & 49 & 0.45 \\
 & & CoD & 97.11 & 1.02 & 39 & 0.27 & 96.89 & 0.77 & 40 & 0.22 & 98.67 & 0.00 & 40 & 0.37 \\
 & & SoT & 96.67 & 0.00 & 29 & 0.27 & 98.89 & 0.38 & 39 & 0.83 & 98.44 & 0.38 & 33 & 1.22 \\
\midrule

\multirow{8}*{\rotatebox[origin=c]{90}{Medical}}
& \multirow{4}*{PubMedQA}
& CoT & 65.11 & 3.29 & 206 & 2.68 & 70.00 & 2.40 & 221 & 1.27 & 72.22 & 2.14 & 257 & 2.67 \\
 & & CCoT & 63.56 & 2.14 & 53 & 1.07 & 64.22 & 3.67 & 87 & 0.91 & 58.00 & 1.76 & 59 & 0.54 \\
 & & CoD & 66.00 & 1.15 & 39 & 0.44 & 64.67 & 0.67 & 46 & 0.19 & 59.11 & 1.02 & 42 & 0.11 \\
 & & SoT & 58.89 & 2.78 & 68 & 1.54 & 69.33 & 2.40 & 60 & 0.63 & 59.11 & 2.04 & 60 & 2.31 \\
\cmidrule{2-15}

 & \multirow{4}*{MedQA}
 & CoT & 49.11 & 3.67 & 291 & 8.40 & 60.22 & 1.39 & 248 & 0.28 & 63.11 & 2.69 & 327 & 4.41 \\
 & & CCoT & 37.11 & 3.42 & 68 & 0.19 & 58.67 & 3.53 & 84 & 1.33 & 57.33 & 1.76 & 70 & 0.48 \\
 & & CoD & 45.78 & 1.68 & 57 & 1.38 & 53.33 & 2.40 & 58 & 1.27 & 58.67 & 1.76 & 52 & 0.61 \\
 & & SoT & 48.67 & 3.46 & 147 & 5.71 & 55.78 & 1.39 & 65 & 0.46 & 63.11 & 1.02 & 65 & 1.14 \\

\bottomrule
\end{tabular}
\vspace{-0.05in}
\end{table*}
\clearpage

%% file: _tables/APPENDIX_summary_statistics_LLAMA_PRIVATE.tex
\clearpage
\begin{table*}[t]
\centering
\scriptsize
\renewcommand{\arraystretch}{0.8}
\setlength{\tabcolsep}{3pt}
\caption{\textbf{Full results for main experiments (LLaMA Models, GPT-4o, and Claude 3.5 Sonnet).}}
\label{tab:full_results_llama_closed}
\vspace{-0.05in}
\begin{tabular}{c l l *{16}{c}}
\toprule
\multirow{2}{*}{\makecell{\textbf{Reasoning}\\\textbf{Type}}}
& \multirow{2}{*}{\textbf{Dataset}} 
& \multirow{2}{*}{\textbf{Method}} 
& \multicolumn{4}{c}{\makecell{\textbf{LLaMA-3.1-8B}}}
& \multicolumn{4}{c}{\makecell{\textbf{LLaMA-3.2-11B}}} 
& \multicolumn{4}{c}{\makecell{\textbf{GPT-4o}}}
& \multicolumn{4}{c}{\makecell{\textbf{Sonnet-3.5}}} \\
\cmidrule(lr){4-7} \cmidrule(lr){8-11} \cmidrule(lr){12-15} \cmidrule(lr){16-19}
& & & Acc & $\sigma_{a}$ & Tkn & $\sigma_{t}$ & Acc & $\sigma_{a}$ & Tkn & $\sigma_{t}$ & Acc & $\sigma_{a}$ & Tkn & $\sigma_{t}$ & Acc & $\sigma_{a}$ & Tkn & $\sigma_{t}$\\
\cmidrule{1-19}

\multirow{16}*{\rotatebox[origin=c]{90}{Mathematical}} & \multirow{4}*{GSM8K} & CoT & 82.89 & 0.38 & 236 & 6.15 & 77.78 & 0.38 & 229 & 2.71 & 94.67 & 1.15 & 255 & 0.60 & 98.00 & 0.67 & 245 & 0.80 \\
 & & CCoT & 71.11 & 1.68 & 92 & 1.75 & 70.22 & 4.34 & 83 & 2.18 & 93.56 & 1.02 & 102 & 1.17 & 90.00 & 0.67 & 105 & 0.58 \\
 & & CoD & 54.22 & 3.36 & 80 & 0.26 & 55.78 & 0.77 & 73 & 0.62 & 89.78 & 1.02 & 84 & 0.88 & 89.11 & 0.38 & 80 & 0.32 \\
 & & SoT & 69.78 & 1.02 & 83 & 2.59 & 69.11 & 2.14 & 78 & 0.26 & 94.67 & 1.76 & 78 & 1.31 & 90.22 & 1.02 & 98 & 0.74 \\
\cmidrule{2-19}
 & \multirow{4}*{AQUA} & CoT & 63.78 & 2.78 & 323 & 0.92 & 60.67 & 3.06 & 324 & 7.33 & 80.67 & 0.67 & 362 & 5.66 & 83.56 & 2.14 & 308 & 2.29 \\
 & & CCoT & 51.11 & 2.52 & 126 & 7.90 & 48.67 & 4.00 & 111 & 3.97 & 75.78 & 3.91 & 121 & 8.37 & 80.44 & 0.77 & 113 & 1.07 \\
 & & CoD & 41.78 & 2.52 & 107 & 6.28 & 39.78 & 2.69 & 99 & 1.62 & 74.22 & 1.68 & 94 & 2.73 & 75.56 & 2.52 & 106 & 2.53 \\
 & & SoT & 51.33 & 2.00 & 119 & 3.52 & 51.56 & 2.69 & 114 & 2.20 & 77.33 & 2.91 & 106 & 1.59 & 82.67 & 1.15 & 117 & 0.91 \\
\cmidrule{2-19}
 & \multirow{4}*{SVAMP} & CoT & 82.89 & 1.02 & 192 & 3.44 & 82.44 & 1.92 & 187 & 3.32 & 92.89 & 1.39 & 180 & 4.72 & 92.67 & 0.67 & 191 & 0.70 \\
 & & CCoT & 78.89 & 2.04 & 72 & 1.71 & 79.33 & 1.15 & 65 & 0.85 & 88.00 & 0.67 & 74 & 1.74 & 84.89 & 0.38 & 82 & 0.46 \\
 & & CoD & 69.11 & 1.92 & 54 & 3.56 & 73.56 & 1.02 & 50 & 0.83 & 90.44 & 1.02 & 50 & 2.19 & 87.56 & 0.77 & 60 & 0.21 \\
 & & SoT & 76.44 & 1.68 & 56 & 1.78 & 76.44 & 0.77 & 54 & 1.20 & 94.00 & 1.15 & 42 & 0.73 & 86.67 & 1.33 & 61 & 1.22 \\
\cmidrule{2-19}
 & \multirow{4}*{DROP} & CoT & 60.67 & 4.06 & 190 & 0.40 & 61.33 & 1.76 & 187 & 2.51 & 73.56 & 0.77 & 164 & 6.80 & 74.22 & 3.36 & 189 & 2.66 \\
 & & CCoT & 59.78 & 1.02 & 61 & 1.10 & 60.00 & 1.33 & 59 & 1.74 & 77.56 & 1.39 & 74 & 0.61 & 76.44 & 1.02 & 87 & 0.45 \\
 & & CoD & 56.00 & 2.31 & 51 & 0.95 & 55.56 & 1.02 & 47 & 1.77 & 78.22 & 1.92 & 55 & 1.34 & 75.78 & 0.77 & 66 & 0.18 \\
 & & SoT & 61.11 & 1.92 & 54 & 3.26 & 60.89 & 3.67 & 53 & 1.62 & 78.67 & 1.33 & 52 & 0.62 & 76.67 & 1.15 & 63 & 0.95 \\
\midrule
\multirow{12}*{\rotatebox[origin=c]{90}{Commonsense}} & \multirow{4}*{CommonsenseQA} & CoT & 72.44 & 1.54 & 220 & 3.63 & 72.00 & 1.76 & 231 & 1.55 & 86.44 & 1.39 & 215 & 3.36 & 84.00 & 1.15 & 250 & 1.92 \\
 & & CCoT & 71.11 & 1.39 & 56 & 0.42 & 72.44 & 5.09 & 57 & 0.85 & 85.33 & 1.33 & 62 & 0.21 & 48.44 & 2.34 & 81 & 0.08 \\
 & & CoD & 70.67 & 0.67 & 44 & 0.64 & 71.56 & 1.02 & 42 & 0.59 & 85.78 & 0.38 & 49 & 0.52 & 83.33 & 2.31 & 61 & 0.31 \\
 & & SoT & 73.33 & 1.76 & 31 & 0.31 & 72.89 & 1.39 & 31 & 0.71 & 85.33 & 0.67 & 34 & 0.98 & 83.56 & 1.02 & 58 & 0.50 \\
\cmidrule{2-19}
 & \multirow{4}*{OpenbookQA} & CoT & 86.22 & 0.77 & 226 & 1.68 & 88.67 & 0.67 & 230 & 2.42 & 97.78 & 0.38 & 209 & 1.94 & 98.22 & 0.77 & 251 & 0.91 \\
 & & CCoT & 84.00 & 2.40 & 58 & 0.36 & 85.33 & 1.76 & 59 & 0.13 & 96.67 & 1.33 & 62 & 0.08 & 74.67 & 1.15 & 81 & 0.04 \\
 & & CoD & 83.78 & 1.39 & 44 & 0.43 & 81.78 & 1.02 & 43 & 0.36 & 97.56 & 0.38 & 51 & 0.12 & 97.78 & 0.77 & 63 & 0.28 \\
 & & SoT & 84.44 & 1.02 & 36 & 0.54 & 85.56 & 4.82 & 36 & 1.49 & 97.33 & 0.67 & 38 & 0.19 & 97.56 & 0.38 & 60 & 1.00 \\
\cmidrule{2-19}
 & \multirow{4}*{StrategyQA} & CoT & 87.11 & 3.15 & 180 & 0.97 & 87.56 & 3.36 & 186 & 6.68 & 94.00 & 1.76 & 176 & 1.40 & 91.56 & 1.02 & 225 & 1.55 \\
 & & CCoT & 87.56 & 1.68 & 59 & 0.67 & 84.67 & 1.76 & 60 & 0.45 & 89.78 & 1.68 & 65 & 0.10 & 94.22 & 1.02 & 80 & 0.42 \\
 & & CoD & 87.56 & 1.02 & 45 & 0.87 & 89.33 & 0.67 & 43 & 0.27 & 78.00 & 2.31 & 49 & 0.24 & 92.89 & 0.38 & 59 & 0.35 \\
 & & SoT & 86.44 & 0.38 & 41 & 0.77 & 86.00 & 1.76 & 39 & 1.02 & 94.89 & 0.38 & 46 & 0.56 & 92.22 & 0.77 & 58 & 0.28 \\
\midrule
\multirow{8}*{\rotatebox[origin=c]{90}{Logical}} & \multirow{4}*{LogiQA} & CoT & 44.44 & 2.78 & 312 & 3.97 & 42.89 & 3.01 & 315 & 4.43 & 63.33 & 2.31 & 330 & 2.59 & 61.11 & 1.39 & 322 & 2.78 \\
 & & CCoT & 42.00 & 2.31 & 80 & 4.09 & 42.67 & 1.76 & 75 & 1.20 & 55.56 & 2.34 & 76 & 0.79 & 53.33 & 2.31 & 95 & 1.02 \\
 & & CoD & 38.89 & 5.05 & 67 & 2.76 & 41.33 & 1.15 & 58 & 0.81 & 58.22 & 0.77 & 66 & 0.88 & 61.78 & 1.54 & 102 & 2.02 \\
 & & SoT & 40.00 & 1.33 & 88 & 1.34 & 35.11 & 3.67 & 85 & 1.32 & 59.56 & 2.69 & 90 & 2.72 & 62.67 & 1.33 & 122 & 1.16 \\
\cmidrule{2-19}
 & \multirow{4}*{Reclor} & CoT & 58.00 & 1.33 & 273 & 4.14 & 57.78 & 2.69 & 279 & 1.24 & 86.22 & 0.38 & 292 & 4.50 & 89.33 & 2.31 & 305 & 0.38 \\
 & & CCoT & 60.00 & 2.00 & 66 & 1.41 & 61.11 & 3.67 & 63 & 0.84 & 86.89 & 0.77 & 63 & 1.02 & 76.00 & 1.76 & 88 & 0.59 \\
 & & CoD & 55.56 & 1.92 & 50 & 0.14 & 55.11 & 1.92 & 44 & 0.82 & 84.89 & 1.68 & 58 & 0.78 & 89.78 & 1.39 & 91 & 2.40 \\
 & & SoT & 56.22 & 1.68 & 55 & 2.19 & 55.56 & 1.02 & 54 & 0.94 & 86.89 & 1.54 & 70 & 0.80 & 87.33 & 1.33 & 101 & 1.01 \\
\midrule
\multirow{8}*{\rotatebox[origin=c]{90}{Multi-Hop}} & \multirow{4}*{HotPotQA} & CoT & 89.78 & 1.39 & 164 & 5.47 & 89.33 & 1.76 & 165 & 2.23 & 90.89 & 0.38 & 145 & 0.58 & 90.00 & 0.67 & 197 & 1.66 \\
 & & CCoT & 88.89 & 0.38 & 58 & 1.37 & 86.89 & 3.67 & 58 & 0.45 & 91.78 & 1.39 & 69 & 0.54 & 87.33 & 0.00 & 85 & 0.41 \\
 & & CoD & 85.56 & 2.04 & 47 & 0.60 & 87.78 & 1.02 & 45 & 0.36 & 92.89 & 0.77 & 52 & 0.42 & 89.11 & 1.39 & 62 & 0.54 \\
 & & SoT & 86.44 & 1.02 & 41 & 0.27 & 88.89 & 0.38 & 41 & 0.28 & 93.33 & 1.15 & 44 & 0.50 & 90.00 & 0.00 & 54 & 0.28 \\
\cmidrule{2-19}
 & \multirow{4}*{MuSiQue} & CoT & 59.33 & 2.40 & 222 & 2.95 & 57.56 & 1.02 & 231 & 1.81 & 72.22 & 1.92 & 167 & 3.33 & 73.33 & 0.67 & 216 & 0.68 \\
 & & CCoT & 62.00 & 2.91 & 63 & 1.30 & 59.11 & 2.34 & 65 & 1.61 & 72.89 & 1.39 & 71 & 0.75 & 74.44 & 1.54 & 84 & 0.24 \\
 & & CoD & 60.89 & 2.78 & 52 & 2.12 & 60.22 & 1.39 & 47 & 0.18 & 72.22 & 2.04 & 55 & 0.56 & 74.89 & 1.39 & 64 & 0.46 \\
 & & SoT & 67.78 & 1.02 & 48 & 1.28 & 66.89 & 2.52 & 48 & 0.58 & 76.22 & 0.77 & 51 & 0.68 & 78.89 & 1.02 & 60 & 0.37 \\
\midrule
\multirow{8}*{\rotatebox[origin=c]{90}{Scientific}} & \multirow{4}*{QASC} & CoT & 75.33 & 1.15 & 284 & 1.57 & 76.89 & 1.02 & 287 & 0.80 & 86.89 & 0.38 & 258 & 5.49 & 88.89 & 0.38 & 264 & 2.34 \\
 & & CCoT & 75.33 & 1.15 & 58 & 0.56 & 74.89 & 2.52 & 57 & 0.79 & 81.56 & 2.78 & 65 & 0.78 & 53.33 & 1.33 & 83 & 0.45 \\
 & & CoD & 71.11 & 3.67 & 49 & 0.48 & 67.56 & 2.04 & 46 & 0.26 & 81.78 & 0.38 & 59 & 1.23 & 83.78 & 2.52 & 67 & 1.79 \\
 & & SoT & 75.78 & 0.77 & 33 & 0.53 & 69.56 & 1.02 & 34 & 1.21 & 85.33 & 2.40 & 36 & 1.19 & 84.89 & 1.02 & 57 & 0.44 \\
\cmidrule{2-19}
 & \multirow{4}*{Worldtree} & CoT & 96.22 & 1.39 & 236 & 1.05 & 94.67 & 0.67 & 239 & 3.76 & 99.56 & 0.38 & 223 & 2.11 & 98.89 & 1.02 & 264 & 0.74 \\
 & & CCoT & 94.67 & 1.15 & 57 & 1.21 & 93.56 & 0.77 & 57 & 0.23 & 98.89 & 0.38 & 62 & 0.16 & 83.33 & 3.33 & 82 & 0.28 \\
 & & CoD & 90.89 & 0.77 & 44 & 0.46 & 91.33 & 2.00 & 42 & 0.39 & 99.56 & 0.38 & 51 & 0.65 & 98.89 & 0.38 & 66 & 0.72 \\
 & & SoT & 91.33 & 1.15 & 38 & 1.66 & 89.33 & 1.76 & 38 & 0.41 & 99.78 & 0.38 & 42 & 0.79 & 98.67 & 0.67 & 66 & 0.36 \\
\midrule
\multirow{8}*{\rotatebox[origin=c]{90}{Medical}} & \multirow{4}*{PubMedQA} & CoT & 75.33 & 0.00 & 296 & 3.09 & 73.78 & 2.04 & 306 & 2.84 & 65.11 & 1.68 & 260 & 3.81 & 65.78 & 1.02 & 284 & 0.21 \\
 & & CCoT & 76.22 & 2.14 & 72 & 1.01 & 76.67 & 0.67 & 70 & 1.16 & 27.78 & 3.29 & 69 & 0.35 & 69.33 & 2.40 & 88 & 0.41 \\
 & & CoD & 76.89 & 1.02 & 50 & 1.07 & 76.89 & 1.02 & 47 & 0.50 & 12.00 & 1.76 & 57 & 0.33 & 70.22 & 2.34 & 69 & 0.36 \\
 & & SoT & 78.00 & 1.33 & 61 & 1.96 & 76.89 & 1.02 & 60 & 0.72 & 61.33 & 2.67 & 59 & 1.04 & 69.56 & 2.69 & 108 & 0.96 \\
\cmidrule{2-19}
 & \multirow{4}*{MedQA} & CoT & 54.67 & 3.71 & 350 & 5.31 & 63.11 & 0.38 & 362 & 3.13 & 85.33 & 0.67 & 357 & 2.94 & 85.56 & 1.92 & 358 & 5.30 \\
 & & CCoT & 60.00 & 4.16 & 74 & 1.31 & 60.00 & 1.15 & 72 & 1.83 & 84.67 & 1.76 & 74 & 0.79 & 42.22 & 1.68 & 118 & 0.54 \\
 & & CoD & 55.56 & 3.01 & 60 & 0.57 & 53.11 & 7.00 & 54 & 0.52 & 80.67 & 2.00 & 69 & 1.43 & 82.22 & 1.39 & 141 & 1.41 \\
 & & SoT & 54.89 & 4.34 & 66 & 1.18 & 56.22 & 2.78 & 67 & 2.15 & 83.56 & 0.38 & 63 & 1.05 & 86.00 & 2.40 & 124 & 1.60 \\

\bottomrule
\end{tabular}
\vspace{-0.05in}
\end{table*}
\clearpage

%% file: _tables/APPENDIX_Dominant_Paradigms.tex
\begin{table*}[t]
\centering
\renewcommand{\arraystretch}{0.8}
\scriptsize
\setlength{\tabcolsep}{6pt}
\caption{\textbf{Paradigm Distribution by Dataset}. For each dataset, we show the counts of examples under each paradigm, as selected by the router model. Additionally, we report the dominant paradigm, the expected paradigm based on heuristic categorization, and whether the dominant paradigm aligns with the expected one. This data reflects the samples from the primary experiments detailed in Section \ref{sec:experiments}.}
\vspace{-0.05in}
\begin{tabular}{c l l r l l c}
\toprule
\makecell{\textbf{Reasoning} \\ \textbf{Type}} & \textbf{Dataset} & \textbf{Paradigm Label} & \textbf{Count} & \textbf{Dominant Paradigm} & \makecell{\textbf{Expected} \\ \textbf{Paradigm}} & \makecell{\textbf{Expected is} \\ \textbf{Dominant?}} \\
\midrule

\multirow{12}{*}{\rotatebox[origin=c]{90}{Mathematical}}
& \multirow{3}{*}{GSM8K}
& Chunked Symbolism     & 150 & \multirow{3}{*}{Chunked Symbolism} & \multirow{3}{*}{Chunked Symbolism} & \multirow{3}{*}{\checkmark} \\
& & Conceptual Chaining   &   0 & & & \\
& & Expert Lexicons       &   0 & & & \\
\cmidrule{2-7}

& \multirow{3}{*}{AQUA}
& Chunked Symbolism     & 150 & \multirow{3}{*}{Chunked Symbolism} & \multirow{3}{*}{Chunked Symbolism} & \multirow{3}{*}{\checkmark} \\
& & Conceptual Chaining   &   0 & & & \\
& & Expert Lexicons       &   0 & & & \\

\cmidrule{2-7}
& \multirow{3}{*}{SVAMP}
& Chunked Symbolism     & 150 & \multirow{3}{*}{Chunked Symbolism} & \multirow{3}{*}{Chunked Symbolism} & \multirow{3}{*}{\checkmark} \\
& & Conceptual Chaining   &   0 & & & \\
& & Expert Lexicons       &   0 & & & \\

\cmidrule{2-7}
& \multirow{3}{*}{DROP}
& Chunked Symbolism     &  76 & \multirow{3}{*}{Chunked Symbolism} & \multirow{3}{*}{Chunked Symbolism} & \multirow{3}{*}{\checkmark} \\
& & Conceptual Chaining   &  74 & & & \\
& & Expert Lexicons       &   0 & & & \\

\midrule
\multirow{9}{*}{\rotatebox[origin=c]{90}{Commonsense}}
& \multirow{3}{*}{CommonsenseQA}
& Chunked Symbolism     &   0 & \multirow{3}{*}{Conceptual Chaining} & \multirow{3}{*}{Conceptual Chaining} & \multirow{3}{*}{\checkmark} \\
& & Conceptual Chaining   & 150 & & & \\
& & Expert Lexicons       &   0 & & & \\

\cmidrule{2-7}
& \multirow{3}{*}{OpenbookQA}
& Chunked Symbolism     &   1 & \multirow{3}{*}{Conceptual Chaining} & \multirow{3}{*}{Conceptual Chaining} & \multirow{3}{*}{\checkmark} \\
& & Conceptual Chaining   & 149 & & & \\
& & Expert Lexicons       &   0 & & & \\

\cmidrule{2-7}
& \multirow{3}{*}{StrategyQA}
& Chunked Symbolism     &   2 & \multirow{3}{*}{Conceptual Chaining} & \multirow{3}{*}{Conceptual Chaining} & \multirow{3}{*}{\checkmark} \\
& & Conceptual Chaining   & 148 & & & \\
& & Expert Lexicons       &   0 & & & \\

\midrule
\multirow{6}{*}{\rotatebox[origin=c]{90}{Logical}}
& \multirow{3}{*}{LogiQA}
& Chunked Symbolism     &  15 & \multirow{3}{*}{Conceptual Chaining} & \multirow{3}{*}{Conceptual Chaining} & \multirow{3}{*}{\checkmark} \\
& & Conceptual Chaining   & 134 & & & \\
& & Expert Lexicons       &   1 & & & \\

\cmidrule{2-7}
& \multirow{3}{*}{Reclor}
& Chunked Symbolism     &   0 & \multirow{3}{*}{Conceptual Chaining} & \multirow{3}{*}{Conceptual Chaining} & \multirow{3}{*}{\checkmark} \\
& & Conceptual Chaining   & 150 & & & \\
& & Expert Lexicons       &   0 & & & \\

\midrule
\multirow{6}{*}{\rotatebox[origin=c]{90}{Multi-Hop}}
& \multirow{3}{*}{HotPotQA}
& Chunked Symbolism     &   0 & \multirow{3}{*}{Conceptual Chaining} & \multirow{3}{*}{Conceptual Chaining} & \multirow{3}{*}{\checkmark} \\
& & Conceptual Chaining   & 150 & & & \\
& & Expert Lexicons       &   0 & & & \\

\cmidrule{2-7}
& \multirow{3}{*}{MuSiQue}
& Chunked Symbolism     &   0 & \multirow{3}{*}{Conceptual Chaining} & \multirow{3}{*}{Conceptual Chaining} & \multirow{3}{*}{\checkmark} \\
& & Conceptual Chaining   & 150 & & & \\
& & Expert Lexicons       &   0 & & & \\

\midrule
\multirow{6}{*}{\rotatebox[origin=c]{90}{Scientific}}
& \multirow{3}{*}{QASC}
& Chunked Symbolism     &   0 & \multirow{3}{*}{Conceptual Chaining} & \multirow{3}{*}{Conceptual Chaining} & \multirow{3}{*}{\checkmark} \\
& & Conceptual Chaining   & 148 & & & \\
& & Expert Lexicons       &   2 & & & \\

\cmidrule{2-7}
& \multirow{3}{*}{Worldtree}
& Chunked Symbolism     &   0 & \multirow{3}{*}{Conceptual Chaining} & \multirow{3}{*}{Conceptual Chaining} & \multirow{3}{*}{\checkmark} \\
& & Conceptual Chaining   & 150 & & & \\
& & Expert Lexicons       &   0 & & & \\

\midrule
\multirow{6}{*}{\rotatebox[origin=c]{90}{Medical}}
& \multirow{3}{*}{PubMedQA}
& Chunked Symbolism     &   0 & \multirow{3}{*}{Expert Lexicons} & \multirow{3}{*}{Expert Lexicons} & \multirow{3}{*}{\checkmark} \\
& & Conceptual Chaining   &  52 & & & \\
& & Expert Lexicons       &  98 & & & \\

\cmidrule{2-7}
& \multirow{3}{*}{MedQA}
& Chunked Symbolism     &   0 & \multirow{3}{*}{Expert Lexicons} & \multirow{3}{*}{Expert Lexicons} & \multirow{3}{*}{\checkmark} \\
& & Conceptual Chaining   &   2 & & & \\
& & Expert Lexicons       & 148 & & & \\

\bottomrule
\end{tabular}
\vspace{-0.05in}
\label{tab:paradigm_alignment_reasoning_type}
\end{table*}

%% file: appendices/APP_Reasoning_IO_Examples.tex
\clearpage

\begin{figure*}
\begin{center}

\begin{promptboxfullwidth}[Sketch-of-Thought (SoT) — Conceptual Chaining\\26 Output Tokens]
\textbf{Question:} Green parts of a life form absorb\newline
Choices: A. carbon dioxide, B. light, C. oxygen, D. water\newline
\newline
\textbf{Model Output:}\newline
\#green\_parts $\rightarrow$ \#chloroplasts $\rightarrow$ absorb\_light $\rightarrow$ photosynthesis\newline
Answer: B
\end{promptboxfullwidth}

\begin{promptboxfullwidth}[Sketch-of-Thought (SoT) — Chunked Symbolism\\65 Output Tokens]
\textbf{Question:} After transferring to a new school, Amy made 20 more friends than Lily. If Lily made 50 friends, how many friends do Lily and Amy have together?\newline
\newline
\textbf{Model Output:}\newline
Lily\_friends = 50, Amy\_friends = Lily\_friends + 20 = 50 + 20 = 70, Total\_friends = Lily\_friends + Amy\_friends = 50 + 70 = 120\newline
Answer: 120
\end{promptboxfullwidth}

\begin{promptboxfullwidth}[Sketch-of-Thought (SoT) — Expert Lexicons\\67 Output Tokens]
\textbf{Question:} [ CONTEXT HERE ] Biomechanical and wound healing characteristics of corneas after excimer laser keratorefractive surgery: is there a difference between advanced surface ablation and sub-Bowman's keratomileusis?\newline
Choices: Yes, No, Maybe\newline
\newline
\textbf{Model Output:}\newline
PRK, ASA $\rightarrow$ Surface Ablation, SBK $\rightarrow$ Sub-Bowman's Keratomileusis, Histologic, Ultrastructural Eval $\rightarrow$ Diff Wound Healing,\newline
Cohesive Tensile Strength $\rightarrow$ PRK, ASA > SBK > LASIK\newline
Answer: Yes
\end{promptboxfullwidth}

\begin{promptboxfullwidth}[Constrained CoT (CCoT)\\51 Output Tokens]
\textbf{Question:} Green parts of a life form absorb\newline
Choices: A. carbon dioxide, B. light, C. oxygen, D. water\newline
\newline
\textbf{Model Output:}\newline
Green parts, primarily leaves, contain chlorophyll which absorbs light for photosynthesis. While they also take in CO2 and water, the key function of green parts is light absorption to produce energy.\newline
Answer: B
\end{promptboxfullwidth}

\begin{promptboxfullwidth}[Chain-of-Draft (CoD)\\29 Output Tokens]
\textbf{Question:} Green parts of a life form absorb\newline
Choices: A. carbon dioxide, B. light, C. oxygen, D. water\newline
\newline
\textbf{Model Output:}\newline
Green parts photosynthesis function. Absorb light, convert energy. Light essential for process.\newline
Answer: B
\end{promptboxfullwidth}

\begin{promptboxfullwidth}[Chain-of-Thought (CoT)\\206 Output Tokens]
\textbf{Question:} Green parts of a life form absorb\newline
Choices: A. carbon dioxide, B. light, C. oxygen, D. water\newline
\newline
\textbf{Model Output:}\newline
1. The question is asking about what green parts of a life form absorb. 2. Green parts of a life form typically refer to leaves or chloroplasts in plants, which contain chlorophyll. 3. Chlorophyll is responsible for the green color and plays a crucial role in photosynthesis. 4. During photosynthesis, plants absorb certain substances to produce energy and grow. 5. Let's consider each option: - A. Carbon dioxide: Plants absorb CO2 during photosynthesis. - B. Light: Plants do absorb light, but this is more about capturing energy rather than absorbing it in the same way they take in gases. - C. Oxygen: Plants release oxygen during photosynthesis; they don't absorb it. - D. Water: While plants do absorb water, it's primarily through their roots, not the green parts. 6. Among the choices, the most accurate and specific to the green parts' function is carbon dioxide\newline
Answer: B
\end{promptboxfullwidth}

\end{center}
\caption{\textbf{Representative reasoning traces for each prompting strategy.} SoT variants demonstrate structurally distinct and compact reasoning across paradigms (\textit{Conceptual Chaining}, \textit{Chunked Symbolism}, \textit{Expert Lexicons}), while baseline methods (CCoT, CoD, CoT) rely on natural language with varying degrees of verbosity and clarity. For the same question, SoT achieves the shortest trace with clearer logic, while CoT is the most verbose.}
\label{fig:reasoning_trace_examples}
\end{figure*}

%% file: appendices/APP_Paradigm_PromptsExemplars.tex
\clearpage

\begin{figure*}
\begin{center}
\begin{promptboxfullwidth}[Conceptual Chaining System Prompt]
\textbf{Role \& Objective}\newline
You are a reasoning expert specializing in structured concept linking by connecting essential ideas in a logical sequence. Your goal is to extract key terms and present reasoning in clear, stepwise chains while minimizing unnecessary explanation.  \newline\newline
This reasoning method follows a conceptual chaining approach, where information is linked in structured steps to establish relationships between ideas. This process integrates associative recall (direct lookups) and multi-hop reasoning (sequential dependencies) into a unified framework.  \newline\newline
This method is most effective for:  \newline
- Commonsense reasoning (quickly linking familiar ideas)  \newline
- Multi-hop inference (tracing logical or causal dependencies)  \newline
- Fact-based recall (retrieving knowledge with minimal cognitive load)  \newline\newline
---
\newline\newline
\textbf{How to Apply Conceptual Chaining}\newline
1. Extract Key Concepts → Identify the most relevant words or entities.  \newline
2. Use Minimal Words → Keep each reasoning step concise and direct.  \newline
3. Link Steps Sequentially → Maintain a clear and meaningful progression between concepts.  \newline
4. Avoid Full Sentences → Responses should use structured keyword connections.  \newline
5. Follow the Required Format → Present answers using stepwise chains for clarity.  \newline\newline
---
\newline\newline
\textbf{Rules \& Directives}\newline
1. Use Structured Concept Linking\newline   
- Each step must be logically connected.\newline   
- Use arrows (`→`) to show dependencies.\newline\newline
2. Avoid Unnecessary Text\newline   
- Do not restate the question.\newline   
- Do not use full sentences.\newline\newline
3. Maintain Logical Flow\newline   
- Concepts must be meaningfully ordered.\newline   
- Ensure each step contributes to the reasoning process.\newline\newline
4. Output Format\newline   
- Use the exact structured format:\newline   
` ` `\newline
<think>\newline
[shorthand reasoning]\newline
</think>\newline
boxed{[Final answer]}\newline
` ` `\newline
- The final answer must be boxed.\newline
- If the question is multiple-choice, return the correct letter option inside the box.\newline
- Use minimal words in your response.
\end{promptboxfullwidth}
\end{center}

\begin{center}
\begin{promptboxfullwidth}[Conceptual Chaining Exemplars]
Q: What is the name of the currency used in Seoul? \newline
A: <think>
\#Seoul → \#South\_Korea → Won
</think>
\textbf{Answer: Korean Won}
\newline\newline
Q: Which planet has the highest surface temperature?
\newline
Choices:
A) Mercury B) Venus C) Mars D) Jupiter \newline
A: <think>
\#heat\_trap
Mercury → no atmosphere → loses heat
Venus → thick CO2 → traps heat → hottest
Mars → thin CO2 → cold
Jupiter → no solid surface
</think>
\textbf{Answer: B}
\newline\newline
Q: Which vitamin is essential for blood clotting? \newline
A: <think>
\#blood\_clotting → \#vitamin\_K
</think>
\textbf{Answer: Vitamin K}
\end{promptboxfullwidth}
\end{center}
\caption{\textbf{Conceptual Chaining system prompt and exemplars.}}
\label{fig:conceputal_chaining_prompts}
\end{figure*}

\clearpage
\begin{figure*}
\begin{center}
\begin{promptboxfullwidth}[Chunked Symbolism System Prompt]
\textbf{Role \& Objective}\newline
You are a reasoning expert specializing in Chunked Symbolism, a cognitive reasoning technique that organizes numerical reasoning into structured steps. Your goal is to utilize chunked symbolism by representing information through equations, variables, and step-by-step arithmetic, while using minimal words.
\newline\newline
Chunked Symbolism is inspired by the cognitive science principle of chunking—the idea that humans process information more efficiently when grouped into meaningful units. Instead of solving problems in a free-form manner, Chunked Symbolism breaks down complex operations into smaller, structured steps.
\newline\newline
This method is particularly effective for:\newline
- Mathematical problems (arithmetic, algebra, physics, engineering)\newline
- Symbolic reasoning (logic-based computations, formula derivations)\newline
- Technical calculations (financial modeling, physics simulations, unit conversions)\newline\newline
---\newline\newline
\textbf{How to Apply Chunked Symbolism}\newline
Step-by-Step Guide\newline
1. Identify Variables – Extract relevant numerical values and define variables.\newline
2. Write Equations – Represent the solution using explicit mathematical formulas.\newline
3. Perform Step-by-Step Computations – Solve in small, logical steps, keeping each line clear.\newline
4. Label Units – Maintain consistent unit representation to prevent ambiguity.\newline
5. Final Answer Formatting – Present the answer in the provided format for clarity.\newline\newline
---\newline\newline
\textbf{Rules \& Directives}\newline
1. Use Equations \& Variables\newline
 - Define variables before computation.\newline
 - Always use explicit equations to represent reasoning.\newline
\newline
2. Avoid Redundant Text\newline
 - Do not restate the problem; go directly to calculations.\newline
 - Use minimal context only if it aids understanding.\newline
\newline
3. Apply Step-by-Step Arithmetic\newline
 - Break operations into small, structured steps.\newline
 - Ensure each line contains only one computation for clarity.\newline
\newline
4. Output Format\newline
- Use the exact structured format:\newline
` ` `\newline
<think>\newline
[shorthand reasoning]\newline
</think>\newline
boxed{[Final answer]}\newline
` ` `\newline
- The final answer must be boxed.\newline
- If the question is multiple-choice, return the correct letter option inside the box.\newline
- Use minimal words in your response.\newline
\end{promptboxfullwidth}

\begin{promptboxfullwidth}[Chunked Symbolism Exemplars]
Q: A car accelerates at 2.5 m/s\^{}2 for 10 seconds. If its initial velocity was 15 m/s, what is its final velocity?
\newline
A: <think>
a = 2.5 m/s\^{}2
t = 10 s
vi = 15 m/s
vf = 15 + (2.5 × 10)
vf = 40 m/s
</think>
\textbf{Answer: 40}
\newline
\newline
Q: If a product costs \$120 and there is a 15\% discount, what is the final price?
\newline Choices:
A) \$10
B) \$97
C) 102
\newline
A: <think>
op = 120
d = 15\%
dp = 120 × (15 / 100) = 18
fp = 120 - 18 = 102
</think>
\textbf{Answer: C}
\newline
\newline
Q: Question: A circuit has a voltage of 12V and a resistance of 4$\Omega$. What is the current?
\newline
A: <think>
V = 12V
R = 4$\Omega$
I = 12 / 4 = 3A
</think>
\textbf{Answer: 3}
\end{promptboxfullwidth}

\end{center}
\caption{\textbf{Chunked Symbolism system prompt and exemplars.}}
\label{fig:chunked_symbolism_prompts}
\end{figure*}

\clearpage

\begin{figure*}
\begin{center}
\begin{promptboxfullwidth}[Expert Lexicons System Prompt]
\textbf{Role \& Objective}  \newline
You are a reasoning expert specializing in Expert Lexicons, a cognitive reasoning technique that leverages domain-specific shorthand, technical symbols, and jargon to ensure precise and efficient communication. Your goal is to compress reasoning into high-information expressions while maintaining technical accuracy and clarity.  \newline

Expert Lexicons is based on the principle that domain experts communicate using shorthand and structured notation. Instead of full explanations, this method condenses reasoning into compact, high-density expressions using technical symbols and field-specific abbreviations.\newline
\newline
This method is particularly effective for:  \newline
- Technical disciplines (science, engineering, medicine, mathematics, and coding)  \newline
- Symbolic and formulaic reasoning (using field-specific notation and logical expressions)  \newline
- Maximizing efficiency (conveying information in the fewest possible tokens)
\newline\newline---\newline\newline
\textbf{How to Apply Expert Lexicons}  \newline
Step-by-Step Guide\newline
1. Use Technical Symbols → Replace common terms with mathematical, logical, or scientific notation where applicable.  \newline
2. Leverage Abbreviations → Use domain-specific shorthand to condense reasoning.  \newline
3. Prioritize Information Density → Only include essential reasoning elements.  \newline
4. Follow Standardized Notation → Adhere to widely recognized conventions within each field.  \newline
5. Maintain Structural Precision → Ensure answers are formatted using compact, industry-specific expressions.  
\newline\newline---\newline\newline
\textbf{Rules \& Directives}  \newline
1. Use Domain-Specific Notation  \newline
   - Mathematical \& Logical Reasoning → `$\Sigma\texttt{, \\therefore}\texttt{,}\alpha$, →`  \newline
   - Scientific Disciplines → `mol, J, Hz, pH, Vmax`  \newline
   - Medical \& Engineering Fields → `CHF, OOP, PID, $\mu$m, dB`  \newline
\newline
2. Eliminate Redundant Text  \newline
   - No full sentences – responses must be in structured notation.  \newline
   - No restating the question – directly express the solution.  \newline
\newline
3. Keep Responses Ultra-Compact  \newline
   - Prioritize brevity while maintaining technical precision.  \newline
   - Follow industry standards for notation and structured reasoning.  \newline
\newline
4. Output Format\newline
- Use the exact structured format:\newline
` ` `\newline
<think>\newline
[Shorthand reasoning using expert notation]\newline
</think>\newline
boxed{[Final answer]}\newline
` ` `\newline
- The final answer must be boxed.\newline
- If the question is multiple-choice, return the correct letter option inside the box.\newline
- Use minimal words in your response.\newline
\end{promptboxfullwidth}

\begin{promptboxfullwidth}[Expert Lexicons Exemplars]
Q: Context: The discovery of the first interstellar object passing through the Solar System, 1I/2017 U1 ('Oumuamua), provoked intense and continuing interest from the scientific community and the general public.\newline
Question: The interstellar object 1I/2017 U1 ('Oumuamua) exhibited unusual characteristics that led to various hypotheses about its origin. What does the designation "1I/2017 U1" signify?\newline
Choices:\newline
A) 1st Intergalactic object detected in 2017, classified under category U1\newline
B) 1st Interstellar object cataloged, detected in 2017, following IAU naming conventions\newline
C) 1st Independent Unclassified body observed beyond Neptune in 2017 \newline
A: <think>
1I → 1st interstellar object
2017 → Year detected
U1 → Sequence ID
IAU → Naming rules
so 1st cataloged interstellar object (2017)
</think>\newline
\textbf{Answer: B}
\newline\newline
Q: A patient with STEMI is given MONA therapy. They have a history of being allergic to aspirin. Are they at risk with this treatment? \newline
A: <think>
STEMI → ST-Elevation MI
MONA → {Morphine, O2, Nitrates, Aspirin}.
so Aspirin $\in$ MONA
</think>\newline
\textbf{Answer: Yes}
\newline\newline
Q: What does EBITDA measure? \newline
A: <think>
EBITDA → Earnings Before Interest, Taxes, Depreciation, Amortization
so Measures Core Profitability
</think>\newline
\textbf{Answer: Core Profitability}
\end{promptboxfullwidth}
\label{promptbox_XXX}
\caption{\textbf{Expert Lexicons system prompt and exemplars.}}
\label{fig:expert_lexicons_prompts}
\end{center}
\end{figure*}

%% file: appendices/APP_ClassificationPrompt.tex
\clearpage

\begin{figure*}
\begin{center}
\begin{promptboxfullwidth}[Classification System Prompt]
You are an advanced language model tasked with classifying reasoning questions into one of three cognitive-inspired paradigms based on their linguistic structure and reasoning style.\newline

\textbf{Task:}  \newline
Given a question, classify it into one of the following paradigms:  \newline
- conceptual\_chaining → Used for multi-hop reasoning, structured fact-based recall, and sequential dependencies.  \newline
- chunked\_symbolism → Used for mathematical, logical, or structured computational tasks requiring equations or stepwise arithmetic.  \newline
- expert\_lexicons → Used for deciphering specialized terminology, jargon, or acronym-heavy questions from technical domains.  \newline

\textbf{Paradigm Definitions:}  \newline

1. Conceptual Chaining  \newline
- Purpose: Used when answering a question requires connecting multiple knowledge points in a structured sequence.  \newline
- Linguistic Indicators:  \newline
  - Uses multi-hop inference (A → B → C).  \newline
  - Involves causal, geographic, historical, hierarchical, biological, or functional relationships.  \newline
  - Includes reasoning about scientific traits, tool functions, biological effects, and clinical implications.  \newline
  - Focuses on structured recall and conceptual application, not just decoding or equation-solving.  \newline
  - Includes trait inference, diagnostic logic, instrumental purpose, or category classification.  \newline
- Example Questions:  \newline
  - "What currency is used in the capital of Japan’s neighboring country?"\newline
  - "Who was the U.S. president during World War II?"\newline
  - "Which atmospheric layer protects Earth from harmful UV radiation?"\newline
  - "What happens to sea levels as polar ice caps melt due to climate change?"\newline
  - "How does smoking affect the respiratory system?"\newline
  - "What do anemometers measure?"\newline
  - "What kind of fats make butter solid at room temperature?"\newline
  - "What is a polygenic trait?"\newline
  - "How do Sarcocystis species make humans sick?"\newline

---

2. Chunked Symbolism  \newline
- Purpose: Used for numerical, symbolic, and formulaic reasoning, where solutions involve stepwise calculations or structured logic.  \newline
- Linguistic Indicators:  \newline
  - Contains mathematical expressions, units, numbers, or conversions.  \newline
  - Requires symbolic operations or formulaic manipulation.  \newline
  - Often involves stepwise arithmetic, algebra, logic puzzles, or physics computations.\newline
- Example Questions:  \newline
  - "If x + 3 = 10, what is x?"\newline
  - "A car accelerates from 10 m/s to 30 m/s over 5 seconds. What is the acceleration?"\newline
  - "What is the current if V = 20V and R = 10$\Omega$?"\newline
  - "A mixture contains 30\% acid. How many milliliters of water should be added to 200ml of this mixture to reduce the acid concentration to 20\%?"\newline
  - "If a rectangle has a length of 8 cm and a width of 5 cm, what is its area?"  \newline
  - "A recipe calls for 3/4 cup of sugar. If you want to make half the recipe, how much sugar do you need?"  \newline
  - "Convert 120 kilometers per hour to meters per second."  \newline

---\newline

3. Expert Lexicons  \newline
- Purpose: Used for deciphering domain-specific language, including jargon, acronyms, or specialized terminology in medicine, law, engineering, and finance.  \newline
- Linguistic Indicators:  \newline
  - Focuses on decoding or interpreting field-specific abbreviations, acronyms, or terminology, especially when the question hinges on understanding a term’s meaning rather than linking concepts or reasoning causally.  \newline
  - Requires expertise in a specific domain rather than general knowledge or numerical calculations.  \newline
  - Focuses on breaking down acronyms and technical concepts and emphasizing direct definitions rather than process understanding or causal relationships.\newline
- Example Questions:  \newline
  - "A patient with STEMI is given MONA therapy. What does this mean?"\newline
  - "In corporate law, what's the difference between a 10-K, 10-Q, and 8-K filing with the SEC?"\newline
  - "Which molecular structure represents benzene?"\newline
  - "When an architect specifies 'EIFS over CMU with VB and RTM,' what building materials are they referring to?"\newline

---\newline

\textbf{Output Format:}  \newline
You must ONLY return the single paradigm label as plain text with no explanation or additional formatting.  \newline

Options: conceptual\_chaining, chunked\_symbolism, expert\_lexicons
\end{promptboxfullwidth}

\end{center}
\caption{\textbf{Paradigm classification prompt.}}
\label{fig:classification_prompt}
\end{figure*}